\documentclass[preprint, 10pt]{elsarticle}
\usepackage{geometry}
\usepackage{lineno,hyperref}
\journal{Applied Soft Computing}
\usepackage{times}
\usepackage{epsfig}
\usepackage{amssymb}
\usepackage{pdflscape} 
\usepackage{bbm}
\usepackage{xcolor}
\usepackage{amsmath} 
\usepackage{multirow,booktabs}
\usepackage{mathtools}
\usepackage{soul}
\usepackage{xcolor}         
\usepackage{bm}             
\usepackage[capitalise]{cleveref}       
\usepackage{graphicx}       
\usepackage{booktabs}       
\usepackage{amsfonts}       
\usepackage{nicefrac}       

\usepackage{microtype}      
\usepackage{subcaption}
\mathchardef\mhyphen="2D 
  
\definecolor{red}{rgb}{0.234, 0.6835, 0.808}
\usepackage{color, colortbl}
\definecolor{Gray}{gray}{0.9}

\usepackage{multirow}
\usepackage{tabularx}
\usepackage[linesnumbered,ruled,vlined]{algorithm2e}
\usepackage{algorithmicx}
\usepackage{etoolbox}
\usepackage{algpseudocode}

\begin{document}

\begin{frontmatter}

\title{Finding and Exploring Promising Search Space for the 0-1 Multidimensional Knapsack Problem}

\author{Jitao Xu}

\author{Hongbo Li}
\ead{lihb905@nenu.edu.cn}
\author{Minghao Yin}

\address{College of Information Science and Technology, Northeast Normal University, Changchun, China.}

\begin{abstract}
The 0-1 Multidimensional Knapsack Problem (MKP) is a classical NP-hard combinatorial optimization problem with many engineering applications. In this paper, we propose a novel algorithm combining evolutionary computation with the exact algorithm to solve the 0-1 MKP. It maintains a set of solutions and utilizes the information from the population to extract good partial assignments.  To find high-quality solutions, an exact algorithm is applied to explore the promising search space specified by the good partial assignments. The new solutions are used to update the population. Thus, the good partial assignments evolve towards a better direction with the improvement of the population. Extensive experimentation with commonly used benchmark sets shows that our algorithm outperforms the state-of-the-art heuristic algorithms, \emph{TPTEA} and \emph{DQPSO}, as well as the commercial solver \emph{CPlex}.  It finds better solutions than the existing algorithms and provides new lower bounds for 10 large and hard instances.

\end{abstract}

\end{frontmatter}

\section{Introduction}

The 0-1 Multidimensional Knapsack Problem (MKP) is a well-known combinatorial optimization problem that has been applied in different practical domains. Given a set $N$ = \{1, 2, ..., $n$\} of items with profits $p_i > 0$ and a set $M$ = \{1, 2, ..., $m$\} of resources with a capacity $b_j > 0$ for each resource. Each item $i$ consumes a given amount of each resource $r_{ij} > 0$. The 0-1 MKP is to select a subset of items that maximize the sum of the profits without exceeding the capacity of each resource. The problem is formally stated as:
\begin{align*}
  &\text{Maximize }\  f = \sum^{n}_{i=1} p_ix_i \\
  &\text{s.t. } \sum^{n}_{i=1}r_{ij}x_i \leq b_j, j \in M =\{1,2,...,m\} \\
  &x_i \in \{0, 1\}, i \in N =\{1,2,...,n\}
\end{align*}

\noindent where each $x_i$ is a binary variable indicating whether the item $i$ is selected, i.e., $x_i$ = 1 if the item is selected, and 0 otherwise.

The 0-1 MKP has many real-world engineering applications, such as resource allocation \cite{rf46, resouce, cloud}, cutting stock \cite{rf1}, portfolio-selection \cite{rf48}, obnoxious facility location\cite{rf49}, slice admission control problem\cite{SAC}, food order optimization problems\cite{DPA}, and multi-unit combinatorial auction\cite{MUCA}. Due to its NP-hard characteristic, solving the MKP is computationally challenging. The exact algorithms for the MKP are usually based on branch and bound(B\&B) algorithms, such as the CORAL algorithm \cite{coral}. However, the modern exact algorithms could solve only the MKP instances of relatively small and moderate size, i.e., the instances with less than 250 items and less than 10 resource constraints. For larger instances with more than 500 items and more than 30 resource constraints, the heuristic algorithms are usually better choices.

There exist a number of heuristic algorithms for the problem. For instance, the single-solution based local search algorithms maintain a solution during the search process and iteratively optimize it. The representative of these algorithms include Tabu Search\cite{rf11}, Simulated Annealing \cite{sa}, Kernel Search\cite{rf13}, and Core Problem Algorithm\cite{Cp}. In recent years,  the population-based evolutionary algorithms are more popular for the problem, such as Particle Swarm Optimization(PSO)\cite{1SACRO, 2SACRO, rf40, rf14, dqpso, chih2023stochastic}.  Ant Colony Optimization\cite{rf21, rf16}, Moth Search Optimization\cite{rf17}, Pigeon Inspired Optimization\cite{rf30, rf31}, Bee Colony Algorithm\cite{rf32}, Quantum Cuckoo Search Algorithm\cite{rf27}, Fruit Fly Optimization Algorithm\cite{rf42, rf44} and Grey Wolf Optimizer\cite{rf39}. \citet{1SACRO} introduced two algorithms based on binary PSO to address the MKP. Building upon this research, \citet{2SACRO} further improved the efficacy of PSO in handling the MKP by incorporating the self-adaptive check and repair operator (SACRO) mechanism. This enhancement was subsequently refined in 2018 by introducing an advanced SACRO concept integrating three pseudo-utility ratios to augment efficiency \cite{rf40}. In a different approach, \citet{rf14} adopts quantum particle swarm optimization (QPSO) to address the MKP, while \citet{dqpso} innovated a diversity-preserving mechanism aimed at enhancing the diversity of the population in QPSO. Hybrid algorithms have also been proposed, such as the NP+BAS+LP algorithm introduced by \citet{rf21}. This hybrid approach integrates Binary Ant System (BAS) within a Nested Partition (NP) framework, leveraging Linear Programming (LP) relaxations to guide partitioning and improve solution quality. BAS facilitates the generation of feasible solutions within each partition, thereby enhancing search efficiency. On the nature-inspired optimization front, \citet{rf17} proposed a binary moth search algorithm specifically designed for the MKP. Additionally, \citet{rf30} introduced Pigeon-Inspired Optimization (PIO) for solving the MKP. Further contributions include an enhanced binary heuristic algorithm based on the Artificial Bee Colony (ABC) method by \citet{rf32}, Binary Fruit Fly Optimization Algorithm (bFOA) proposed by \citet{rf44}, and an enhanced version of Fruit Fly Optimization Algorithm (IFFOA) by \citet{rf42}. Each of these algorithms brings unique strategies to the table, such as utilizing binary strings to represent solutions, incorporating various search processes, and employing repair operators to ensure solution feasibility.  Moreover, \citet{rf39} utilized The Binary Grey Wolf Optimizer (GWO) for solving the MKP. This approach combines critical elements, including an initial elite population generator, a quick repair operator based on pseudo-utility, and a novel evolutionary mechanism featuring a differentiated position updating strategy, to achieve improved performance in solving the MKP. Recently, machine learning assisted methods\cite{rf33, rf34, rf35}, along with probability learning-based algorithms\cite{rf29} emerged. \citet{rf33}  tested different machine learning methods to predict each item of the knapsack and select the items that have a higher score predicted by the model. \citet{rf27} combines nearest K-neighbors (KNN) and cuckoo search. They use KNN to improve the diversification of cuckoo search. \citet{rf35} adopted db-scan to perform the binarization process on solutions generated by swarm intelligence algorithms. Although these algorithms provide a new perspective, they are less efficient than the heuristic algorithms when solving large MKP instances. Therefore, heuristics algorithms are still a popular choice when solving some large and hard MKP instances in practice.  Among these heuristic algorithms, the \emph{TPTEA}\cite{rf18} algorithm and  \emph{DQPSO} \cite{dqpso} algorithm can be considered as the state of the art for the MKP.  \emph{TPTEA} incorporates tabu search techniques into a population-based evolutionary framework, creating a two-phase search algorithm that guarantees both effective intensification and diversification across the search space. To our best knowledge, the \emph{TPTEA} algorithm provided the last updates of the lower bounds of the commonly used benchmark set containing the large instances with 500 items and 30 resource constraints. \emph{DQPSO} integrates a diversity-preserving mechanism to ensure a robust variety within the quantum particle swarm, preventing premature convergence of the algorithm. Compared with \emph{TPTEA}, the \emph{DQPSO} algorithm loses a little in the final solution quality, but it finds high-quality solutions earlier than \emph{TPTEA}.

The population-based evolutionary algorithms have good prospects in solving the 0-1 Multidimensional Knapsack Problem. In this paper, we propose a novel algorithm for the 0-1 MKP. The algorithm finds promising partial assignments by an adapted evolutionary algorithm and explores the promising search space specified by the partial assignments with an exact algorithm.  It utilizes the advantages of Evolutionary Computation and Large Neighbourhood Search (\emph{LNS}) \cite{lns} in its major components. It eliminates those search spaces when the exact algorithm is searching, both items that are likely to be selected and unlikely to be chosen. During each search of the exact algorithm, we keep a solution population to avoid wastage. Consequently, it performs well when solving some large and hard 0-1 MKP benchmark instances. Our experiments run with 281 commonly used benchmark instances show that our algorithm performs better than the state-of-the-art heuristic algorithms including \emph{TPTEA} and \emph{DQPSO}.    Our contribution can be summarized as,  (1) We propose to combine Evolutionary Computation with exact algorithm to solve the 0-1 MKP. (2) We adopt and adapt the \emph{TPTEA} algorithm to find promising partial assignments from the population. (3) Our algorithm provides new a lower bound for 8 hard and large instances.

The paper is organized as follows.  We first introduce the motivation and the raw idea of our algorithm in Section 2.  The details of the proposed algorithm, including the framework of the new algorithm,  and the adaption of the \emph{TPTEA} algorithm for finding good partial assignments, a customized \emph{LNS}, are present in Section 3.  The experimental results are shown in Section 4.  We discussed our algorithm with some closely related methods in Section 5.  Finally, Section 6 is the conclusion.

\section{The Motivation}

An \emph{assignment} $\mathcal{A}$ to items $I \subseteq N$ is a set of \emph{instantiations} of the form ($x_i = v_i$), one for each $i \in N$ to assign $v_i$ to $x_i$ where $v_i \in \{0, 1\}$. If $I = N$, then $\mathcal{A}$ is a \emph{complete assignment}; otherwise a \emph{partial assignment}. A complete assignment $s$ that satisfies all resource constraints is a feasible solutions to the instance, or a \emph{solution} for short. A solution $s$ is an \emph{optimal solution} if its objective value $f(s)$ is larger than or equal to that of any other solution to the instance. A partial assignment is an optimal one if it can be extended to an optimal solution.

Given an MKP instance $P$ with $n$ items, the size of the search space of finding an optimal solution for $P$ is $2^n$. Given a partial assignment $\mathcal{A}$  of $P$. Fixing the values of $\mathcal{A}$ in $P$ will result in a sub-problem $P^{\prime}$ whose search space is $2^{n- \vert \mathcal{A} \vert}$. The search space of $P^{\prime}$ can be considered as a sub search space of $P$ which is specified by $\mathcal{A}$. If we are given an optimal partial assignment $\mathcal{A^*}$ of $P$, then we can find an optimal solution for $P$ in a search space of size $2^{n- \vert \mathcal{A^*} \vert}$ instead of the entire search space of size $2^n$. However, as far as we know, there is no existing method that identifies optimal partial assignment for general 0-1 MKP before solving it exactly. Thus, the aim of our algorithm is to find good partial assignments that have a large possibility of being an optimal one, or can be extended to a high-quality solution whose objective value is close to that of an optimal solution. Then exploring the promising search space specified by the good partial assignments may find high-quality solutions efficiently.

Evolutionary Algorithms (\emph{EA}) are powerful techniques for solving combinatorial optimization problems based on the simulation of biological evolution. They start from a population of individuals (representing solutions) and produces offspring (new solutions) by utilizing the information of the population.  Guided by a fitness function measuring the quality of a solution, the new solutions with high-quality are usually used to update the population. As a result, the population evolves towards the direction of better fitness function.  One of the advantages of \emph{EA} is to identify and keep some high-quality patterns (partial assignments) in the population, so they can produce high-quality solutions through crossover and mutation operations.  Thus, we propose a novel algorithm that combines Evolutionary Algorithm with exact algorithm for the 0-1 MKP.   The main idea of the algorithm is shown in Algorithm \ref{alg:idea}.

\begin{algorithm}[!h]
\caption{The main idea}
\label{alg:idea}
\KwIn{a MKP instance $P$;}
\KwOut{current best solution $s^*$;}

$population \leftarrow$ \emph{initPopulation}($P$)\;
$s^* \leftarrow$ the best solution in $population$\;
\While{the termination condition is not reached}
{
    $pa \leftarrow extract(population)$\;
    $s \leftarrow explore(P, pa)$\;
    \emph{updatePopulation}($s$)\;
    \If{$f(s) > f(s^*)$}
    {
        $s^* \leftarrow s$\;
    }
}
\textbf{return} $s^*$\;
\end{algorithm}

At the beginning, the population of Evolutionary Algorithm, a set of solutions recorded in $population$ is initialized at line 1. The $s^*$ records the best solution found so far.  The loop at lines 3-8 is the main procedure, e.g., a promising partial assignment \emph{pa} is extracted from the population  by an Evolutionary Algorithm at line 4 and the promising search space is explored by an exact algorithm at line 5. The new solution is used to update the population at line 6. If a better solution is found, $s^*$ will be updated accordingly at lines 7-8,   The procedure repeats until a termination condition is reached and the best found solution is returned.  There are various termination conditions that can be used in the algorithm, such as a time limit or the number of iterations.

\section{Related Work }

There exist some methods combining Evolutionary Computation with exact algorithms for solving combinatorial optimization problems. We refer those ones related to our method in the following, and then discuss the differences at the end of next section.

Kernel Search \citet{rf13} can be applied in solving the MKP.  It divides the item set into two parts, the kernel, and the buckets, and then applies an exact algorithm in the kernel.  Kernel Search finds the solution of the continuous relaxation problem, and sorts the items according to the information provided by the continuous solution. Then it selects the first \textit{C} items as a kernel $\Lambda$. The items belonging to $N\setminus \Lambda$ construct a sequence $\{B_i\}$ of buckets. Finally, an exact algorithm is iteratively applied in the kernel.  In each iteration, some of the items in buckets are selected and added into the kernel. There are two additional constraints in each run of the exact algorithm: (1) the lower bound is the current best solution; (2) at least one of the latest added items should be selected in the solution.

The Generate And Solve framework (GAS) \cite{gs1,gs2} was originally designed for solving the container loading problem. The framework suggests decomposing the original problem into several sub-instances whose solutions are also the solutions of the original one, and then applying some exact algorithms in the sub-instances.  Besides the GAS framework, the Construct, Merge, Solve $\&$ Adapt algorithm (CMSA) \cite{cmsa} suggests to generates sub-instances by merging different solution components found in probabilistically constructed solutions. The algorithm has been applied in solving the minimum common string partition (MCSP) problem, and a minimum covering arborescence (MCA) problem.

\citet{05ee} propose a hybrid algorithm that combines exact algorithm with evolutionary computation for the MKP.  It runs an evolutionary algorithm and an exact algorithm alternately. The two algorithm components share a single solution. Whenever one algorithm component \emph{A} finds a new solution better than the current best solution, the other algorithm component \emph{B} takes over and the new solution is used to assist algorithm component \emph{B}.  More specifically, when the exact algorithm finds a better solution, the solution is used to replace the worst one in the population of the evolutionary computation. On the other hand, when the evolutionary algorithm finds a better solution, the objective of the solution is used to update the lower bound of the exact algorithm.

\section{Combining Evolutionary Algorithm with Integer Programming for the MKP}

Based on the idea  proposed in last section, we introduce the detailed algorithm for solving the 0-1 MKP, namely \emph{finding and exploring promising search space} (\emph{FEPSS}).  The framework of the algorithm is presented first.

\subsection{The Framework}

The main procedure of \emph{FEPSS} is present in Algorithm \ref{alg:frame}.  At the beginning, the population, a set of solutions, is initialized at line 1 and the current best solution is initialized to the best one of the initial population. There are two cases to find and explore good partial assignments.  In the first case (lines 10-12), the good partial assignments are extracted from the population at line 11 and the search space specified by $pa$ is explored at line 12.   In the second case (lines 6-8), we run a customized Large Neighbourhood Search procedure to refine current best solution.  
Each of the two cases returns a new solution $s$, which is used to update the population at line 13 and update the current best solution at lines 14-16. A counter \emph{num} is used to record the number of the first case is executed.  If the counter reaches a predefined parameter \emph{lnsLimit} or the best solution is updated, the second case will be executed once.  The details of the procedures called in the framework will be introduced in the following subsections.

\begin{algorithm}[!h]
\caption{The framework}
\label{alg:frame}
\KwIn{a MKP instance $P$;}
\KwOut{current best solution $s^*$;}

$population \leftarrow$ \emph{initPopulation}($P$)\;
$s^* \leftarrow$ the best solution in $population$\;
$num \leftarrow 0$\;
\While{the termination condition is not reached}
{

    \If{iterationNum = lnsLimit \textbf{or} bestUpdated}
    {
        $bestUpdated \leftarrow false$\;
        $num  \leftarrow 0$\;
        $s \leftarrow$ \emph{LNS($P$, $s^*$)}\;
    }
    \Else{
       $num  \leftarrow num + 1$\;
       $pa \leftarrow$ \emph{extractFromPopulation($P$)}\;
       $s \leftarrow $ \emph{explore}($P$, $pa$)\;
    } 
    \emph{updatePopulation}($s$)\;
    \If{ $f(s) > f(s^*)$}
    {
        $bestUpdated \leftarrow true$\;
        $s^* \leftarrow s$\;
    }
}
\textbf{return} $s^*$\;
\end{algorithm}

Note that there is a preprocessing step ahead of the algorithm, e.g., all the items are preprocessed and renumbered in an ascending order of the surrogate relaxation ratio\cite{ratio} evaluation of the items, which is defined as follows.
\begin{flalign}
    c_i = \frac{p_i}{\sum^{m}_{j=1}\frac{r_{ij}}{b_j}}, \forall i \in \{1,2,...,n\}
\label{f1}
\end{flalign}

After the items are renumbered, the corresponding $p_i$ and $r_{ij}$ are adjusted accordingly.  In the following, all the items will be used with their new numbering. An item $i$ is selected (not selected) in a solution means that the value of the corresponding variable $x_i$ is 1 (0),  and vice versa. $population[k]$ is the $k$-$th$ solution in $population$ and $s[i]$ is the value of variable $x_i$ in solution $s$.

In the following subsections, we will introduce the \emph{initPopulation}, \emph{extractFromPopulation}, \emph{LNS}, \emph{explore} and \emph{updatePopulation} procedures in details.

\subsection{Initializing the Population}

The solution set $population$ contains $ns$ solutions. At the beginning, we randomly generate $ns \times 1000$ solutions and add the best $ns$ ones into $population$ (selected by their objectives). Then we run a tabu search to refine the random solutions in $population$.  The procedure of generating initial $population$ is present in Algorithm \ref{alg:init}.

When generating a random solution, the \emph{geneRandomSolutions} procedure randomly selects an item and tries to add it into the knapsack.  The item is added into the knapsack if no resource constraint is violated; otherwise, it is skipped. A random solution will be generated after all items are tried.

To refine the random solutions, the tabu search strategy of the first phase of \emph{TPTEA} algorithm is employed \cite{rf18}. We briefly recall the procedure here. It explores only the feasible search space and uses two basic neighborhoods: the one-flip neighborhood $N_1(s)$ and the swap neighborhood $N_2(s)$. The $N_1(s)$ contains all the feasible solutions which can be obtained by applying the $flip$ operator that changes the value of the variable $x_q$ in $s$ to its complementary value 1 - $x_q$.  The $N_2(s)$ contains all the feasible solutions which can be obtained by applying the $swap$ operator that swaps the values of two variables with different values in $s$. The \emph{tabuSearch} procedure is shown in Algorithm \ref{alg:tabusearch}.  It always finds the best neighbour from current neighbourhood $N_1(s)\bigcup N_2(s)$ and mains a record of all the visited solutions to avoid duplicate visiting. The tabu function is implemented with three hashing vectors \cite{hash}. The best solution visited during the tabu search is returned.

\begin{algorithm}[!h]
\caption{\emph{initPopulation}}
\label{alg:init}
\KwIn{a MKP instance $P$;}
\KwOut{a solution set $population$;}

$population \leftarrow$ \emph{geneRandomSolutions}()\;
\For{$i=1$ to $ns$} 
{
    $population[i] \leftarrow$ \emph{tabuSearch}($population[i]$)\;
}
\textbf{return} $population$\;
\end{algorithm}

\begin{algorithm}[!h]
\caption{\emph{tabuSearch}}
\label{alg:tabusearch}
\KwIn{a solution $s$;}
\KwOut{a refined solution $s^{\prime}$;}
$iteration \leftarrow 1$\; 
$tabuList \leftarrow \emptyset$\;
$s^{\prime} \leftarrow s$\;
\While{$iteration \leq maxIteration$}
{
	find a best neighbor solution $s_{neighbor}$ from $N_1(s)\bigcup N_2(s)$, which is not in $tabuList$\;
	$s \leftarrow s_{neighbor}$\;
	\If{$f(s) > f(s^{\prime})$}
	{
		$s^{\prime} \leftarrow s$\;
	}
    add $s$ into $tabuList$\;
	$iteration \leftarrow iteration + 1$\;
}

\textbf{return} $s^{\prime}$\;
\end{algorithm}

\subsection{Extracting High-quality Partial Assignments from the Population}

In this subsection, we introduce how to extract good partial assignments from the population.  The idea is to divide the items into two parts: the fixed part and the free part. Then we generate a partial assignment for the fixed part and explore the free part with an exact algorithm. The \emph{extractFromPopulation} procedure is shown in Algorithm \ref{alg:fsss}. It decides items in the fixed part at lines 1-8 and then assigns values to the items in the fixed part at lines 13-25 to generate a partial assignment.  The details of the algorithm are as follows.

\begin{algorithm}[!h]
\caption{\emph{extractFromPopulation}}
\label{alg:fsss}
\KwIn{a MKP instance $P$;}
\KwOut{a partial assignment $pa$;}
calculate $\delta_i$ for each item $i$\;
$t \leftarrow $ number of items selected in $s^*$\;
sort all items by the ascending order of $\delta_i$\;
$fix \leftarrow \emptyset$\;
\For{each $i \in [1, n-t-\Delta)$}
{
    $fix \leftarrow fix \cup \{i\}$\;
}
\For{each  $i \in (n-t+\Delta, n]$}
{
    $fix \leftarrow fix \cup \{i\}$\;
}

$num \leftarrow \frac{ns}{10}$\;
$pa \leftarrow \emptyset$\;
$exceed \leftarrow true$\;
\While{$exceed$}
{

\For{ each $i \in fix$}
{
    $selectedNum \leftarrow 0$\;
    \For{$j=1$ to $num$}
    {
        $s \leftarrow $ select a solution from $population$ randomly\;
        \If{s$[i] = 1$}
        {
            $selectedNum \leftarrow selectedNum + 1$\;
        }
    }
    $v \leftarrow 1$\;
    \If{$selectedNum < \frac{num}{2}$}
    {
        $v \leftarrow 0$\;
    }
    $r \leftarrow rand(0,1)$\;
    \If{$r < \frac{1}{n \times 10}$}
    {
        $v \leftarrow 1 - v$\;
    }
    $pa \leftarrow pa \cup  (x_i, v)$\;
}
\If{the items selected in $pa$ violate no resource constraint}
{
$exceed \leftarrow false$\;
}
\Else{
$pa \leftarrow \emptyset$\;
}
}

\textbf{return} $pa$\;
\end{algorithm}

Firstly, each item $i$ is associated with a score $\delta_i$ defined as:

\begin{flalign}
    \delta_i = \beta\times \frac{w_i}{ns} + (1-\beta)\times\frac{\frac{i}{n}+rand(0,1)}{2}
\label{f2}
\end{flalign}

\noindent where $\beta \in (0, 1)$ is a parameter, $rand(0,1)$ returns a pseudo random number between (0, 1) and $w_i$ is defined as:

\begin{flalign}
   w_i = \sum^{ns}_{k=1} population[k][i]
\end{flalign}

Thus, $\frac{w_i}{ns}$ indicates the proportion of solutions in $population$ selecting item $i$.  The items are renumbered by the score defined in formula (\ref{f1}), so $\frac{i}{n}$ indicates how likely item $i$ is selected by the surrogate relaxation ratio.  We would prefer adding the items with higher $\delta_i$ into the knapsack.

\begin{figure}[!h]
	\centering
	\centerline{\includegraphics[scale=0.5]{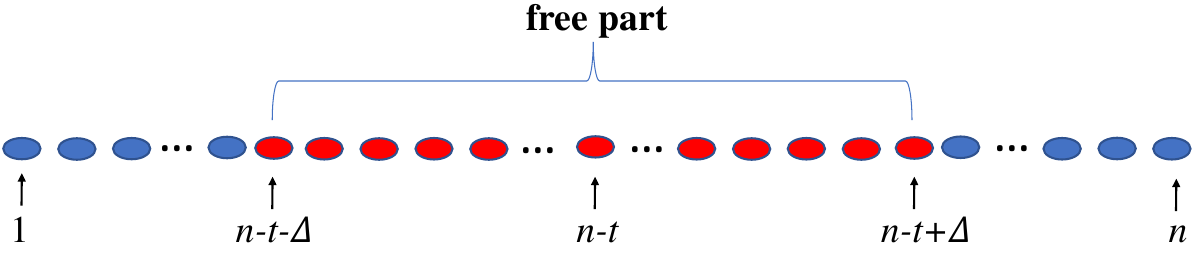}}
	\caption{Example of fixed part and free part.}
	\label{fig:figure1}
\end{figure}

Secondly, we count the number of items selected in current best solution $s^*$, i.e., $t = \sum^{n}_{i=1}s^*[i] $. Then we sort the items in ascending order of $\delta_i$. The items $i \in [1, n-t-\Delta)$ and the items $i \in (n-t+\Delta, n]$ compose a fixed part which is recorded in $fix$ at lines 4-8, and the remaining items  $i \in [n-t-\Delta, n-t+\Delta]$ compose a free part, where $\Delta$ is a parameter. We make such selections because the items with smaller $\delta_i$ have larger possibility of not being selected and the ones with larger $\delta_i$ have larger possibility of being selected, and the free part should contain those hard to decide items. The strategy eliminates the items locating near the head (items with small $\delta_i$) and the end (items with large $\delta_i$) of the ordering. As is shown in Figure \ref{fig:figure1}, the blue ones are fixed part and the red ones are free part. 

Although the score $\delta_i$ gives a recommendation to the values of variables in the fixed part, we design a more elaborate strategy to determine the values to avoid local optimum (lines 14-21 in Algorithm \ref{alg:fsss}). For each item in the fixed part, we randomly select $\frac{ns}{10}$ solutions from the solution set $population$. Then we count the number of the selected solutions containing the item, recorded in $selectedNum$ (lines 15-18). An item is selected if more than half of the $\frac{ns}{10}$ solutions contain the item (lines 19-21). A mutation operation that flips the value with probability $\frac{1}{n\times10}$ is used at lines 22 to 24. The while loop at line 12 and the if condition at line 26 ensure that the generated partial assignment violates no resource constraint. After the partial assignment $pa$ is generated, an exact algorithm (Integer Programming) is employed to explore the corresponding search space, which will be introduced later.

\subsection{The Customized Large Neighbourhood Search}

In this subsection, we introduce a customized Large Neighbourhood Search procedure to refine  current best solution.  The procedure of extracts good partial assignments from current best solution and explores the sub-space specified by the partial assignments to find high-quality solutions. It also uses the information of the population to select some items that compose a free part and the remaining items compose the fixed part. We keep the values in current best solution for the fixed part to generate a partial assignment.  The LNS procedure is shown in Algorithm \ref{alg:cbs}.

\begin{algorithm}[!h]
\caption{\emph{LNS}}
\label{alg:cbs}
\KwIn{a MKP $P$ and current best solution $s^*$}
\KwOut{a solution $s$;}
$s \leftarrow null$\;
\For{$i=1$ to $lnsIterationNum$}
{
    $var_1 \leftarrow$ \emph{votingFunction}()\;
    $var_2 \leftarrow$ \emph{randomFunction}()\;
    $var_3 \leftarrow$ \emph{ratioFunction}()\;
    $free \leftarrow var_1 \cup var_2 \cup var_3$\;
    $pa \leftarrow s^* - free$\;

    $s^{\prime} \leftarrow$ \emph{explore}($P$, $pa$)\;
    \emph{updatePopulation}($s^{\prime}$)\;
    \If{$s=null$ \textbf{or} $f(s^{\prime}) > f(s) $} 
    {
       $s \leftarrow s^{\prime}$\;
    }
}
\textbf{return} $s$\;
\end{algorithm}

We propose three functions, \emph{votingFunction}, \emph{randomFunction} and \emph{ratioFunction} to select the items belonging to the free part. Each of them will find a set of items and their union constitutes the free part (lines 3-7). Then we remove the assignments of the free part from current best solution $s^*$, and the remaining constitute a partial assignment $pa$ (line 7).

The \emph{votingFunction} is based on the score $w_i$ defined in formula (3) which records the number of solutions in $population$ voting for the selection of the item. The function selects items into free part in two cases: (1) item $i$ is in  $s^{*}$ and $w_i < ns-\sigma$; (2) item $i$ is not in  $s^{*}$ and $w_i > \sigma$, where $\sigma$ is the standard deviation of $W = \{w_1, w_2, ..., w_n\}$.

The following two functions work on the renumbered items which are sorted by the ascending order of formula (1). The \emph{randomFunction} simply selects each item indexed from $index_1$ to $n$ with probability $P_1$, where $index_1$ points to the first item with $w_i \ge \frac{ns}{2}$.

The \emph{ratioFunction} uses another index $index_2$ that points to the first item with $w_i \ge 1$. It makes selections in the items indexed from $index_2$ to $index_1$ only. For each item $i$, we calculates its average profit $\frac{p_i}{r_{ij}}$ in each resource constraint $j$. In each resource constraint, we can find a set of items containing the top $t_2$ items with the largest average profit in this constraint. If an item has appeared in any of the sets, it will be selected with a probability $P_2$. 

The \emph{LNS} procedure calls the three functions to select the items belonging to the free part at lines 3-5. After that, it combines the three item sets to get the free part at line 6. Then it generates a partial assignment $pa$ by removing the assignments of the free part from current best solution $s^*$ at line 7. Finally, it calls the \emph{explore} procedure to explore the search space specified by the partial assignment at line 8. The procedure repeats \emph{lnsIterationNum} times and the best solution found in the \emph{LNS} procedure is returned.

\subsection{Exploring A Promising Search Space}

In \emph{explore} procedure shown in Algorithm \ref{alg:ip},  we add all the items selected by the partial assignment into the knapsack and calculate the remaining resource for each resource constraint $j$, $b_j^{\prime}=b_j - \sum_{i \in fixed\ part} r_{ij}x_i$. Then we calculate the total cost, $tc_j=\sum_{i \in free\ part} r_{ij}$, of each resource $j$ of all the free items. The resource constraint with largest deficit is the one with the largest $\frac{tc_j}{b_j^{\prime}}$, marked by $l$ at line 8.  We calculate $\frac{p_i}{b_{l}}$ for each item $i$ involved in $pa$ and sort the items by the descending order of $\frac{p_i}{b_{l}}$. Then the assignment of top $t_1$ items are removed from $pa$, i.e., added into free part. Finally, we fix the values of the items that are still in $pa$ and run Integer Programming to solve the sub problem within a given time limit $T$ seconds, i.e., the IP may not complete. The best solution found within $T$ seconds is returned.

\begin{algorithm}[!h]
\caption{\emph{explore}}
\label{alg:ip}
\KwIn{a MKP instance $P$ and the fixed part $pa$;}
\KwOut{a solution  $s$;}

\For{each resource constraint $j$}
{
    $b_j^{\prime} \leftarrow b_j$\;
    \For{each $(x_i, v) \in pa$}
    {
        $b_j^{\prime} \leftarrow b_j^{\prime} - r_{ij} \times v$\;    
    }
    $tc_j \leftarrow 0$\;
    \For{each item $i$ not involved in $pa$}
    {
        $tc_j \leftarrow tc_j + r_{ij}$\;    
    }
}
$l \leftarrow $ the index of the constraint with largest $\frac{tc_j}{b_j^{\prime}}$\;

calculate $\frac{p_i}{b_{l}}$ for each item involved in $pa$\;
sort the items by the descending order of $\frac{p_i}{b_{l}}$ and
remove the assignment of the top $t_1$ items from $pa$\;

\For{each $(x_i, v) \in pa$}
{
    fix the value of $x_i$ to $v$\;
}

$s \leftarrow $ run Integer Programming to solve the sub problem with the fixed values in $T$ seconds\;
\textbf{return} $s$\;
\end{algorithm}

\subsection{Updating the Population}

When updating the $population$ with a new solution $s$, the \emph{updatePopulation} procedure adds $s$ into $population$, and removes the one with the smallest scoring function $V(s)$ defined as: 
\begin{equation}
\begin{aligned}
V(s) &= \alpha \times \frac{f(s)-f_{min}}{f_{max}-f_{min}}+\frac{1-\alpha}{2} \times   (\frac{D(s)-D_{min}}{D_{max}-D_{min}}+\frac{R(s)-R_{min}}{R_{max}-R_{min}})
\end{aligned}
\end{equation}
\begin{flalign}
&D(s) = Min \{ distance(s, s^{\prime}) : s, s^{\prime} \in population, s \neq s^{\prime} \}
\end{flalign}
\begin{flalign}
	R(s) = \sum^{n}_{i=0} c_i \times s[i]
\end{flalign}

\noindent where $\alpha \in [0-1]$ is a parameter, $distance$ is the Hamming distance, $c_i$ is defined in formula (1), the $min$ and $max$ denote the minimum and maximum ones of the corresponding scores respectively. We always keep current best solution in $population$. If current best solution has the smallest $V(s)$, then we remove the one with the second smallest $V(s)$. Note that if $s$ is already in $population$, the \emph{population} will not updated.

\subsection{Discussions}

Our algorithm can be considered as the combination of Evolutionary Computation \cite{hec} and Large Neighbourhood Search \cite{lns}. It maintains a set of solutions, extracts information from the population to help find high-quality solutions, and updates the population when new solutions are found. All of these operations are within the framework of Evolutionary Computation and are customized for the 0-1 MKP in our algorithm. The idea of LNS is to find a fixed part from the current best solution and explore the remaining free part by other algorithms, so our \emph{LNS} procedure is a specific case of LNS, which extracts information from the population to determine the fixed part of the current best solution and employs Integer Programming to explore the remaining search space. The \emph{extractFromPopulation} procedure is simulating LNS. The major difference is that the fixed part is generated from the population, not from the current best solution. Theoretically, the fixed part extracted by \emph{extractFromPopulation} procedure may exceed some resource constraints, but we did not observe this case in our experiments. On the other hand, the fixed part generated from the current best solution will not exceed any resource constraint. We discuss the differences between \emph{FEPPS} and the related methods in the following.

The main idea of Kernel Search is similar to that of \emph{FEPSS} with the kernel corresponding to the free part of \emph{FEPSS} and the buckets corresponding to the fixed part. However, the procedure of finding the kernel is different from the procedure of finding the free part in \emph{FEPSS}. In addition, the fixed part of \emph{FEPSS} contains both selected items and non-selected ones, while the buckets contain only non-selected items. 

Our algorithm can be considered as an instantiation of the GAS framework \cite{gs1,gs2}.  While the GAS framework suggests decomposing the original problem into several sub-instances, our method employs the idea of evolutionary computation to generate the fixed part (some of the variables are fixed) which can be considered as a special case of sub-instances (the domain of some variables are reduced). The major difference between CMSA and \emph{FEPSS} is also the way of generating sub-instances. In addition, we propose a new scoring function for updating the solution set with the new solutions found by Integer Programming.  

The hybrid algorithm \cite{05ee} runs an evolutionary algorithm and an exact algorithm alternately, so it is different from \emph{FEPSS}, since \emph{FEPSS} utilizes an evolutionary algorithm to find promising search spaces and runs an exact algorithm to explore the promising search spaces.

\section{Experiments and Analysis}

To examine the efficiency of \emph{FEPSS} algorithm, finding high-quality solutions in a reasonable time, we performed vast experimentation with two benchmark sets that are widely used to test the algorithms for the MKP. Before presenting the results, we introduce some detailed settings of the experiments in the following.

\textbf{Benchmark. }
We perform extensive computational experiments on the commonly used benchmark instances, the OR-Library instances, and the MK\_GK instances. The main characteristics of the benchmarks are as follows.

\emph{OR-Library:} These instances were proposed in \cite{rf26}. In this set, the number of items $n$  are 100, 250, and 500. The number of resource constraints $m$ are 5, 10, and 30. Each combination ($n, m$) consists of 30 instances. For each pair ($n, m$), the coefficients ($b_i$) are set using $b_i = \omega \Sigma^n_{j=1} r_{ij}$, where $\omega$ represents a tightness ratio. Specifically, $\omega = 0.25$ for the first ten problems (instance ids from 0 to 9),  $\omega = 0.5$ for the next ten problems (instance ids from 10 to 19), and $\omega = 0.75$ for the remaining ten problems (instance ids from 20 to 29). Here, $r_{ij}$ are integer numbers drawn from a discrete uniform generator U(0, 1000). This is the most commonly used benchmark for the MKP. As far as we know, the optimal solutions for the small instances are proven and the (250, 30), (500, 10), and (500, 30) are still open. The instances are available at \url{http://people.brunel.ac.uk/~mastjjb/jeb/orlib/mknapinfo.html}.

\emph{MK\_GK:}  There are 11 instances with $n \in$ \{100, 200, 500, 1000, 1500, 2500\} and $m \in$ \{15, 25, 50, 100\}. This set contains some very large instances, e.g., the one with 2500 items and 100 resource constraints.  The instances are made available by the authors of \cite{rf18} at \url{https://leria-info.univ-angers.fr/~jinkao.hao/mkp.html}.

 \textbf{Baselines. }
We compared our algorithm with the two state-of-the-art heuristic algorithms, the \emph{TPTEA} algorithm and the \emph{DQPSO} algorithm.  Some of the best-known results are found by the algorithms combining linear programming with tabu search \cite{rf28,hyb05}. The results were obtained by running the algorithms for several days, so we did not involve these algorithms in our experiments.  We have used \emph{CPlex} as the Integer Programming solver. Although IP in \emph{FEPSS} is allocated a short time (10 or 20 seconds) and we are not utilizing the full power of \emph{CPlex}, people are interested in the performance of \emph{CPlex}. Running \emph{CPlex} usually needs a large amount of space, so we include the best results of running \emph{CPlex} with 5Gb RAM for indicative purposes.

 \textbf{Implementation Details. }
We use the source codes of \emph{DQPSO} provided by the authors and implemented a \emph{TPTEA} algorithm because the source codes are not publicly available. Our algorithm is implemented in C++ and compiled using the g++ compiler with the -O3 option \footnote{The source code of our algorithm is available at \url{https://github.com/jetou/FEPSS}}. We conducted experiments on Intel E7-4820 v4 (2.0GHz) running on a Linux system.   The performance of the compared algorithms is measured by solution quality. We also include the CPU time of finding the best solution for indicative purposes.

\textbf{Metrices.}
For each instance, we run each algorithm with 30 random seeds ranging from 1 to 30. Consequently, we obtain the best objective value $f_{best}$ among the 30 runs, while $f_{worst}$ represents the worst objective value among these 30 runs. Additionally, we calculate the average objective value $f_{avg}$ and the average time cost $t_{avg}$ of finding the best solution over the 30 runs. The best one in each comparison is highlighted in bold.  Avg. denotes the average values of each column, and $\#Best$ indicates the number of problems where the corresponding algorithm gets the best performance. The average time cost of finding the best solution in 30 runs is also presented in the following tables.

In the following, we first analyze how different components of \emph{FEPSS} affect the efficiency, and then examine the parameters for \emph{FEPSS} and find some appropriate values for the parameters. Finally, we compare  \emph{FEPSS} with the state-of-the-art heuristic algorithms for the MKP, \emph{TPTEA} and \emph{DQPSO}.

\subsection{Ablation Analysis}

We tested the impact of various components of \emph{FEPSS}, including the customized large neighbourhood search (CLNS), the scoring function $V(s)$, and the score of each item $\delta_i$.  The results are shown in Table \ref{tab:Ablation}. The different versions of \emph{FEPSS} are introduced as follows:

\textbf{\emph{FEPSS-1}}: This is the \emph{FEPSS} without the customized large neighbourhood search. Specifically, we remove the lines 5 to 9 from Algorithm 2.

\textbf{\emph{FEPSS-2}}: This is the version where the scoring function $V(s)$ is removed. Without the scoring function, we randomly select and remove a solution from the population.

\textbf{\emph{FEPSS-3}}: In this version, we use the surrogate relaxation ratio, as referenced in formula (\ref{f1}), to replace the score $\delta_i$ presented in formula (\ref{f2}).

\begin{table*}[!htb]
\centering
\scriptsize\setlength{\tabcolsep}{0.5pt}
\begin{tabular*}{\linewidth}{c@{\extracolsep{\fill}}c*{12}{c} }
\hline
\multirow{2}*{Instance}& \multicolumn{3}{c}{FEPSS} 
&\multicolumn{3}{c}{\emph{FEPSS-1}}
&\multicolumn{3}{c}{\emph{FEPSS-2}}
&\multicolumn{3}{c}{\emph{FEPSS-3}}\\ 
	\cline{2-4} \cline{5-7} \cline{8-10} \cline{11-13}
	\noalign{\smallskip}
	 & {$f_{best}$} & {$f_{avg}$} & {$t_{avg}$} & {$f_{best}$} & {$f_{avg}$} & {$t_{avg}$} & {$f_{best}$} & {$f_{avg}$} & {$t_{avg}$} & {$f_{best}$} & {$f_{avg}$} & {$t_{avg}$}\\ \hline
	500-30-0 & \textbf{116014}  & \textbf{115983.7} & 3978.3 & \textbf{116014} & 115980.5 & \textbf{3382.3} & \textbf{116014} & 115977.1  &4114.5 & \textbf{116014} & 115858.7 & 5085.3 \\ 
	
	500-30-1 & \textbf{114810} & \textbf{114767.3}  & 3818.5 & \textbf{114810} & 114763.7 & \textbf{3528.5} & \textbf{114810} & 114764.3  &  3637.3 & 114769& 114647.9& 4744.4\\ 
	
	500-30-2 & \textbf{116741} & \textbf{116689.5} & \textbf{2915.7} & \textbf{116741} & 116687.9 & 3175.2 & \textbf{116741} & 116689.2  & 3894.8 & 116573&116454.5& 5665.4 \\
	
	500-30-3 & \textbf{115370} & 115291.5 & 4022.8 & \textbf{115370} & \textbf{115291.6} & \textbf{3401.7} & \textbf{115370} & 115288.0  &  3689.5 &115313&115187.7& 5274.8\\
	
	500-30-4  &\textbf{116539} & \textbf{116493.5} & \textbf{4285.1} & 116525 & 116485.5 & 4391.4 & 116525 & 116487.5  & 4571.7 & 116413& 116315.3& 4782.7 \\
	
	500-30-5 & \textbf{115741} &115736.3 & \textbf{3610.7} & \textbf{115741} & \textbf{115736.8} & 3856.2 & \textbf{115741} & 115731.2  &4360.9 & 115705&115597.6& 5016.0 \\ 
	
	500-30-6 & \textbf{114181} &114133.4  & 3556.5 & \textbf{114181} & \textbf{114145.0} & \textbf{3301.7} & \textbf{114181} & 114133.0  &3701.1 & 114017& 113906.4&5084.8 \\ 
	
	500-30-7 & \textbf{114344} & \textbf{114314.4} & \textbf{3800.0} & \textbf{114344} & 114307.5 & 4122.9 & \textbf{114344} & 114304.5  & 3885.8 & 114302&114190.1& 5048.8 \\ 
	
	500-30-8 & \textbf{115419} & \textbf{115419.0} & \textbf{996.0} & \textbf{115419} & \textbf{115419.0} & 1086.4 & \textbf{115419} & \textbf{115419.0}  & 1325.3 & 115319& 115250.8& 4847.9 \\
	
	500-30-9 & \textbf{117116} & 117105.7 & 2210.8 & \textbf{117116} & 117107.2 &  \textbf{2191.5} & \textbf{117116} & \textbf{117109.2}  & 3301.9 & 117104& 116925.0& 5573.8 \\ 
	
	500-30-10 & \textbf{218104} & \textbf{218076.1} & \textbf{2601.1} & \textbf{218104} & 218072.9 & 3107.9 & \textbf{218104} & 218074.2  & 3817.0 & 218042& 217975.8& 5217.0 \\ 
	
	500-30-11 & \textbf{214645} & 214631.7 & 1808.9 & \textbf{214645} & 214630.7 & \textbf{1701.3} & \textbf{214645} & \textbf{214633.0}  &1932.0 &\textbf{214645}& 214581.0& 4690.7\\ 
	
	500-30-12 & 215945 &215925.2 & 3830.4 & \textbf{215978} & 215925.4 & 3415.7 & 215942 & \textbf{215927.2} & \textbf{3055.4} & 215905& 215852.9& 3982.3 \\ 
	
	500-30-13 & \textbf{217910} & \textbf{217865.6}  & \textbf{2595.8} & 217885 & 217864.3 & 2811.6 & 217885 & 217864.4  & 3046.4 & 217836& 217748.3& 5460.5 \\ 
	
	500-30-14 & 215640 & 215636.0  & \textbf{3199.5} & 215640 & 215634.5 & 3605.0 & \textbf{215680} & \textbf{215638.9}  & 4105.6 & 215639& 215584.0& 4423.5\\ 
		
	500-30-15 & \textbf{215919} &\textbf{215877.5} & \textbf{3803.2} & \textbf{215919} & 215876.7 & 4123.5 & \textbf{215919} & 215871.5  & 4794.3 &215837& 215773.3& 5398.7 \\ 
	
	500-30-16 & \textbf{215907} &215884.3 & \textbf{1109.1} & \textbf{215907} & \textbf{215884.6} & 1341.8 & \textbf{215907} & 215883.8  & 1236.8 & 215871&215784.5& 4620.0\\ 
	
	500-30-17 & \textbf{216542}&216480.0  & 3363.9 & \textbf{216542} & 216485.1 & \textbf{3243.3} & 216530 & \textbf{216494.7} & 3569.8 &216450& 216336.2& 5136.2 \\ 
	
	500-30-18 & \textbf{217364} & \textbf{217336.6} & \textbf{4307.9} & \textbf{217364} & \textbf{217336.6} & 4425.9 & \textbf{217364} &  217329.2  & 4919.9 & 217312& 217227.3& 4546.3 \\ 
	
	500-30-19 & \textbf{214739} &214704.2  & 3834.4 & \textbf{214739} & \textbf{214706.6} & \textbf{3330.8}& \textbf{214739} &  214702.4 & 3572.3 &214689& 214596.5&4121.6 \\ 
	
	500-30-20 & \textbf{301675} & \textbf{301666.0}  & 3227.9 & \textbf{301675} & 301663.5 & \textbf{2959.8} & \textbf{301675} & 301664.3  &3367.3 & 301667&301616.3&4670.4 \\ 
		
	500-30-21 & \textbf{300055} &\textbf{300052.5} & \textbf{2272.7} & \textbf{300055} & 300050.7 & 3649.5 & \textbf{300055} & 290050.8 & 2413.4 & \textbf{300055}&299990.7&3712.9 \\
	
	500-30-22 & \textbf{305087} &\textbf{305067.0}& \textbf{3034.7} & \textbf{305087} & 305064.0 & 3455.4 & \textbf{305087} & 305066.1  &3409.5 & 305080&304973.1& 3725.2\\ 
	
	500-30-23 & \textbf{302032} &\textbf{302006.2} & 2839.7 & 302008 & 302004.4 & \textbf{2684.6} & 302015 &  302004.2 &3714.7 &301982&301944.0& 3705.5 \\ 
	
	500-30-24 & \textbf{304447}&304425.1  & 3672.7 & \textbf{304447} & \textbf{304430.7} & \textbf{3317.6} & \textbf{304447} & 304425.0 &3578.3 & 304416&304369.9& 3989.7 \\
	
	500-30-25 & \textbf{297012} &296979.6  & 3537.1 & \textbf{297012} & \textbf{296979.7} & 3362.7 & \textbf{297012} & 296975.7 & \textbf{2816.7} & 296958&296936.5&3457.0 \\ 
	
	500-30-26 & \textbf{303364}&303332.0  & 3626.4 & \textbf{303364} & 303335.2 & 3020.0 & \textbf{303364} & \textbf{303340.2} &\textbf{2660.7} & 303335& 303309.1& 3137.3\\ 
	
	500-30-27 & \textbf{307007} &306994.5& 3111.5 & 306999 & 306994.9 & \textbf{2663.4} & 306999 & \textbf{306995.4}  &  3117.8 & 306935& 306857.4& 4837.2 \\ 
	
	500-30-28 & \textbf{303199} &303177.5  & 2876.6 & \textbf{303199} & \textbf{303178.3} & \textbf{2720.7} & \textbf{303199} & 303177.4 & 2749.8 &303178&303136.0& 4226.2 \\ 
	
	500-30-29 & \textbf{300572} &300536.4 & 3362.7 & 300542 & 300534.9 & 3655.2 & \textbf{300572} & \textbf{300536.7} & \textbf{3284.5} &300543& 300480.0& 4144.3\\ \hline
	
	Avg. & \textbf{211448.0}  & \textbf{211419.6} & 3173.7 & 211445.7 & 211419.3 & \textbf{3167.8} & 211446.7 & 211085.3  & 3388.2 & 211396.8 &211313.6& 4610.9\\
	\#Best & \textbf{28} & \textbf{14} & \textbf{13} & 24 & {10} & 13 & 24 & 9  & 4 & 3 & 0&0 \\ \hline	

\end{tabular*}
\caption{\label{tab:Ablation}Ablation analysis of the (500, 30)  OR-Library instances.}
\end{table*}

Our results indicate that \emph{FEPSS} achieves the best outcomes in terms of $f_{best}$ and $f_{avg}$. If we remove the CLNS component, \emph{FEPSS-1} saves a little time to find the current best solution and loses a little in solution quality. Moreover, when we replace the population update scoring function with a random strategy, we observe an increase in time consumption and the loss in solution quality, especially evident in the significant drop observed in $f_{avg}$. Additionally, \emph{FEPSS-3} analysis highlights the significant impact of each item's score $\delta_i$ on the performance of \emph{FEPSS}.  \emph{FEPSS-3} loses in both solution quality and time costs.  Based on these observations, we can infer that the effectiveness order of \emph{FEPSS} components follows CLNS, the scoring function $V(s)$, and the score of each item $\delta_i$, in ascending order. The scoring function $\delta_i$ is effective in identifying the hard-to-decide items, so it finds promising search spaces containing high-quality solutions. These findings provide valuable insights into optimizing and refining the \emph{FEPSS} algorithm for enhanced performance and efficiency.

\subsection{Parameter Settings and Analysis}

The OR-Library instances were generated with different tightness, so we examine the parameters in the representations of the instances with different tightness, e.g. 500-30-5 (tightness=0.25), 500-30-15 (tightness=0.5), and 500-30-25 (tightness=0.75). Each value of the parameters are tested with 30 random seeds from 1 to 30.

\begin{figure}
\centering
\begin{subfigure}{0.65\textwidth}
    \includegraphics[width=\textwidth]{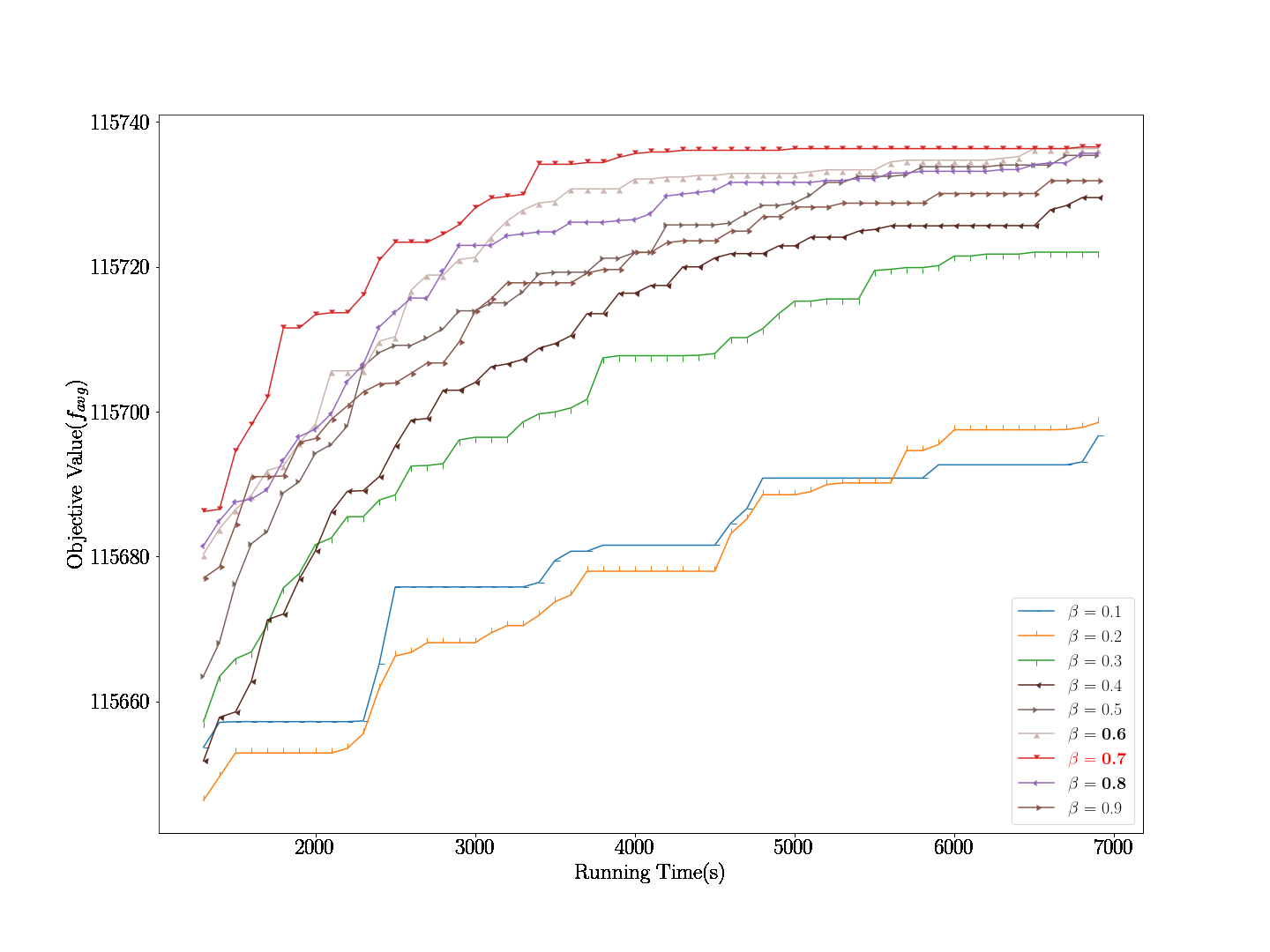}
    \caption{Parameter $\beta$ for 500-30-05(tightness=0.25).}
    \label{fig:sub-beta5}
\end{subfigure}
\hfill
\begin{subfigure}{0.65\textwidth}
    \includegraphics[width=\textwidth]{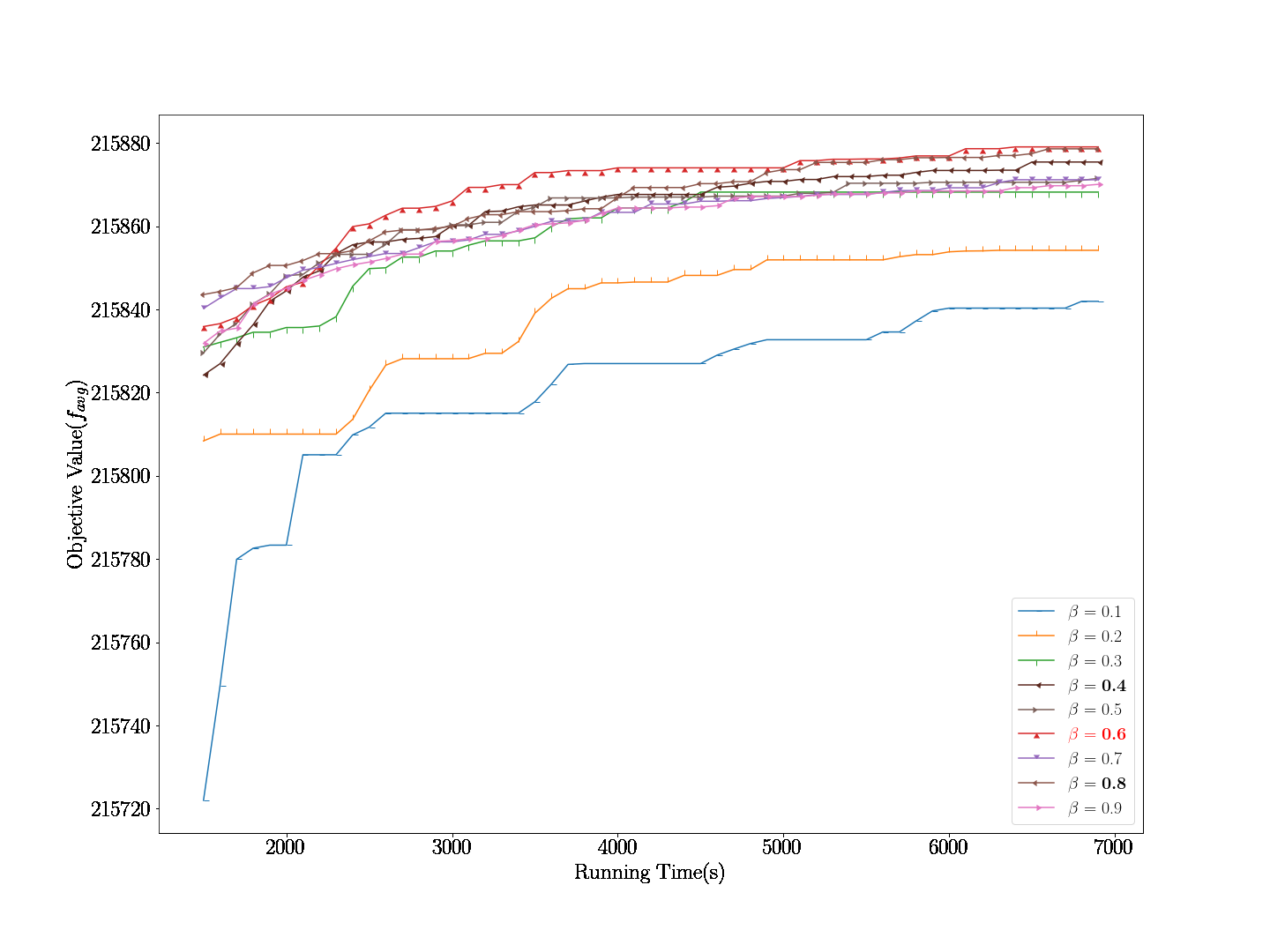}
    \caption{Parameter $\beta$ for 500-30-15(tightness=0.50).}
    \label{fig:sub-beta15}
\end{subfigure}
\hfill
\begin{subfigure}{0.65\textwidth}
    \includegraphics[width=\textwidth]{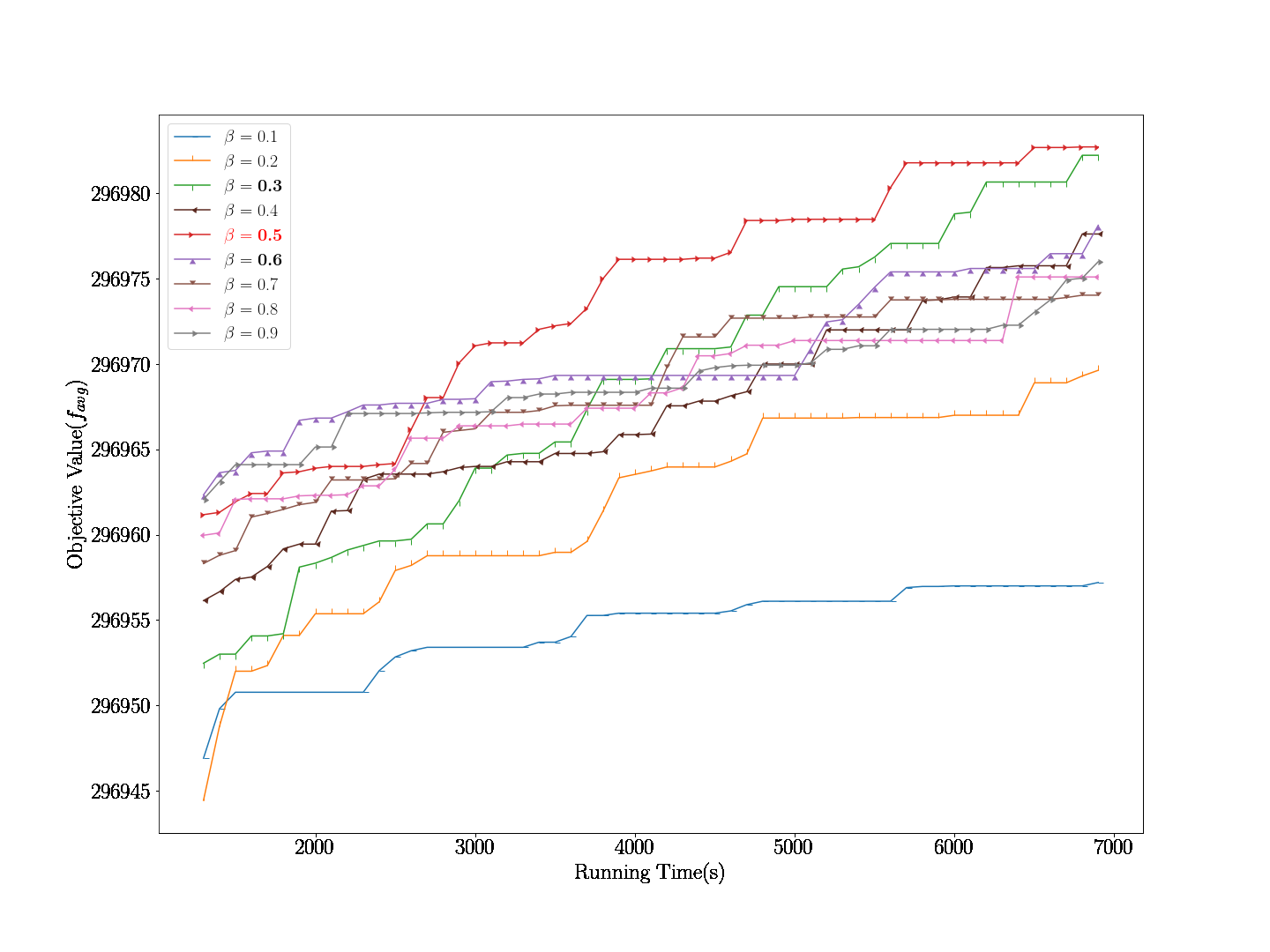}
    \caption{Parameter $\beta$ for 500-30-25(tightness=0.75).}
    \label{fig:sub-beta25}
\end{subfigure}
        
\caption{Influence of the parameter $\beta$ for the average objective value}
\label{fig:Parametersbeta}
\end{figure}

From the ablation study, we observe that the scoring function $\delta_i$ has a large impact on the performance of \emph{FEPSS}, which indicates that the selection of the free part should be further investigated.  Thus, we first examine the parameters affecting the selection of the free part, $\beta$ in formula (\ref{f2}) and $\Delta$ in Figure \ref{fig:figure1}.   We tested $\beta$ ranging from 0.1 to 0.9, and present the results of the three instances in Figure \ref{fig:Parametersbeta}.  The horizontal axis represents the running time, while the vertical axis denotes $f_{avg}$. The legends for the best three of $f_{avg}$ are highlighted in bold and the best one is in red.  We can see that the best values of $\beta$ for the instances with tightness 0.25, 0.5, and 0.75 are 0.7, 0.6, and 0.5 respectively. The average of the best three $\beta$ values are 0.7, 0.6, and 0.47 respectively. The results indicate that with the increasing of the tightness, the best value for $\beta$ should be decreased.  We can also observe that the value 0.6 appears in the best three values for all tightness.  Thus, we use $\beta$=0.6 in the following experiments.

\begin{figure}
\centering
\begin{subfigure}{0.65\textwidth}
    \includegraphics[width=\textwidth]{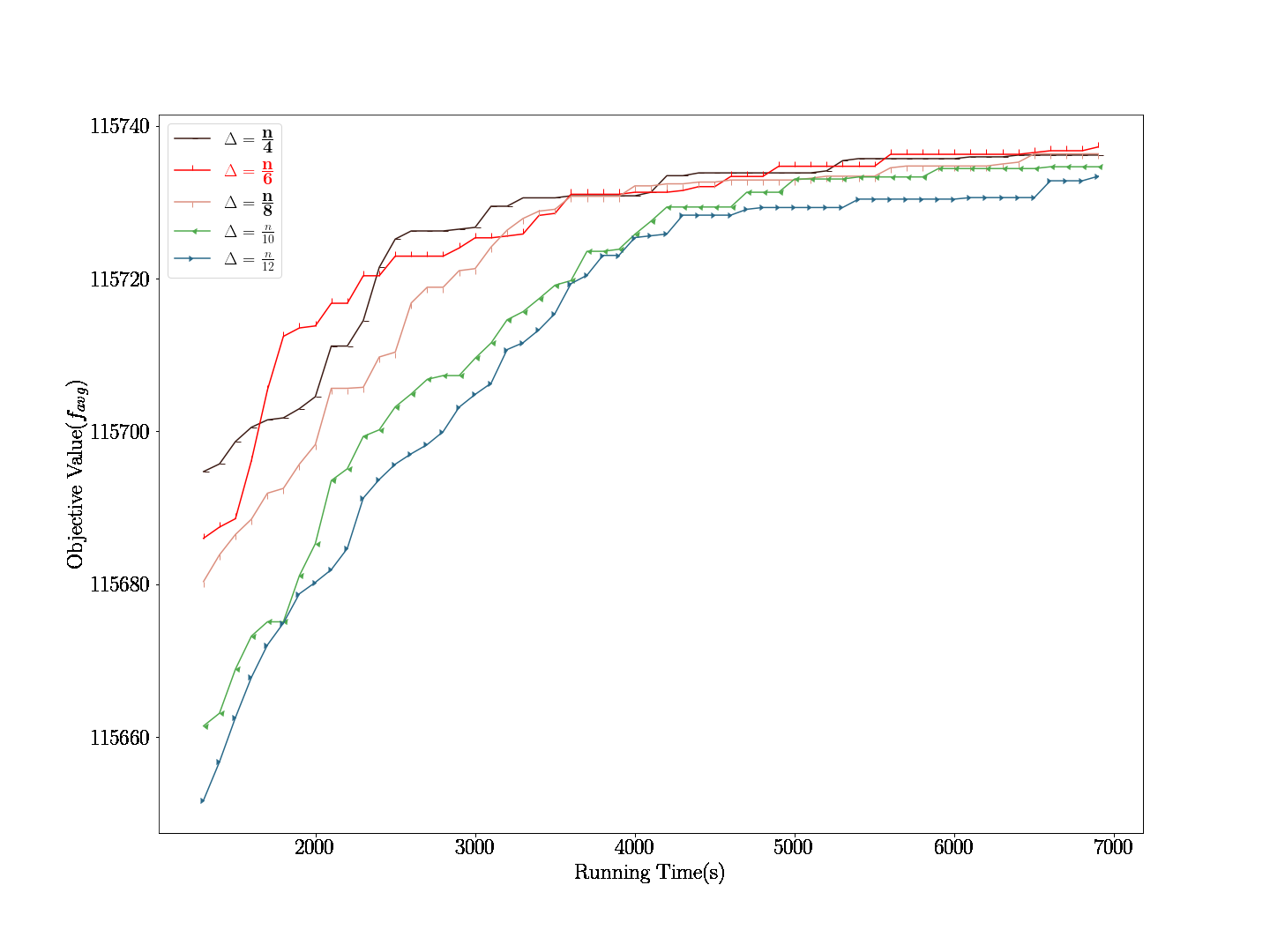}
    \caption{Parameter $\Delta$ for 500-30-05(tightness=0.25).}
    \label{fig:sub-delta5}
\end{subfigure}
\hfill
\begin{subfigure}{0.65\textwidth}
    \includegraphics[width=\textwidth]{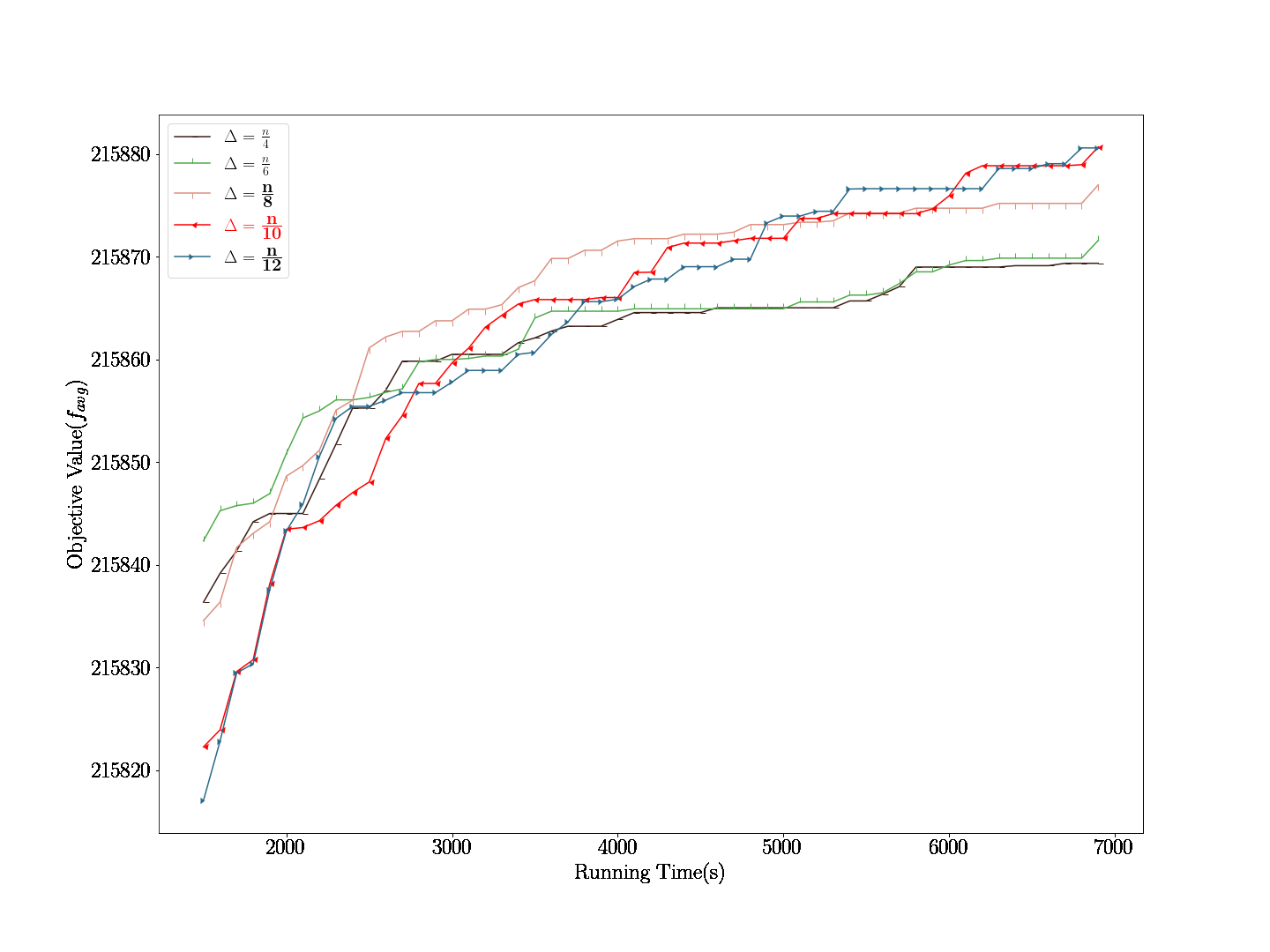}
    \caption{Parameter $\Delta$ for 500-30-15(tightness=0.50).}
    \label{fig:sub-delta15}
\end{subfigure}
\hfill
\begin{subfigure}{0.65\textwidth}
    \includegraphics[width=\textwidth]{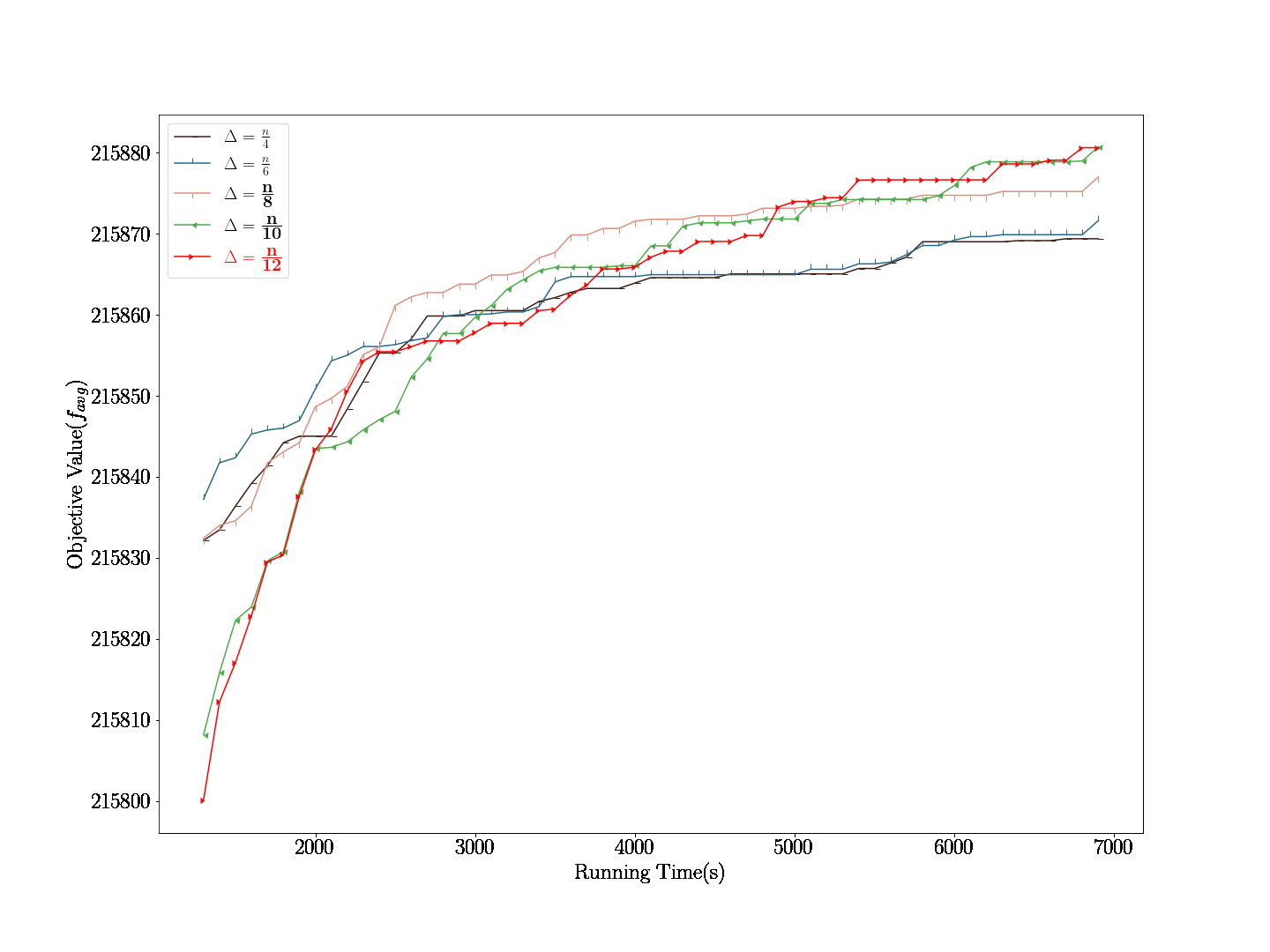}
    \caption{Parameter $\Delta$ for 500-30-25(tightness=0.75).}
    \label{fig:sub-delta25}
\end{subfigure}
        
\caption{Influence of the parameter $\Delta$ for the average objective value}
\label{fig:Parametersdelta}
\end{figure}

In Figure \ref{fig:Parametersdelta}, we present the results of running \emph{FEPSS} with different values of the parameter $\Delta$, which determines the length of the free part.  We tested the values in $\{\frac{n}{4},\frac{n}{6},\frac{n}{8},\frac{n}{10},\frac{n}{12}\}$. The horizontal axis represents the running time, while the vertical axis denotes $f_{avg}$. The legends corresponding to the top three $f_{avg}$ values are highlighted in bold and the best one is in red in red.  In the figure, we observe a similar phenomenon as in Figure \ref{fig:Parametersbeta}.  With the increasing of tightness, the best value for $\Delta$ decreases from $\frac{n}{4}$ to $\frac{n}{12}$, and the average of the best three values decrease from $\frac{n}{6}$ to $\frac{n}{10}$.  Thus, we should run \emph{FEPSS} with a smaller $\Delta$ for an MKP instance with larger tightness.  The value $\frac{n}{8}$ appears in the best three values for all tightness. In the following experiments, we use $\Delta$ = $\frac{n}{8}$.

Finally, we examined the parameter \emph{T} that determines the time limit for each run of the \emph{explore} procedure. In other words, \emph{T} is the time limit of each run of the exact algorithm.  We tested \emph{FEPSS} with \emph{T} in \{5, 10, 30, 60, 300\}. Table \ref{tab:T} presents the results.   The underlined results in the $f_{best}$ columns are those better than the best known results. We can see that when setting \emph{T} to 5, 10, 30, 60, and 300, \emph{FEPSS} finds new solutions better than the best known results in 2, 3, 2, 3, and 0 instances.   \emph{FEPSS} with \emph{T} = 10, 30, 60 find the best solution in 25 instances, which is the largest number.  If we further check the average of $f_{best}$, \emph{FEPSS} with \emph{T} = 10 outperforms all the others.  When \emph{T} is set to 300, \emph{FEPSS} performs poorly, so each run of the exact algorithm should not be allocated a large time limit.  This is reasonable because the more time limit is allocated to one run of the exact algorithm, the less promising search spaces are explored.  We also checked the ratio of time spent by the evolutionary algorithm versus the exact algorithm in each run of \emph{FEPSS}. The result is as we expected, e.g, more than 97\% of time(more than 7000 seconds out of 7200 seconds) are spent by the exact algorithm and the evolutionary algorithm spends less than 3\% of time.  In general, \emph{FEPSS} with \emph{T}=10 is suggested and we have found new lower bounds for 5 instances.

\begin{table*}[!htb]
\centering
\scriptsize\setlength{\tabcolsep}{0.5pt}
\begin{tabular*}{\linewidth}{c@{\extracolsep{\fill}}c*{15}{c} }
\hline
\multirow{2}*{Instance}
&\multicolumn{3}{c}{$T=5$}
& \multicolumn{3}{c}{FEPSS($T=10$)} 
&\multicolumn{3}{c}{$T=30$}
&\multicolumn{3}{c}{$T=60$}
&\multicolumn{3}{c}{$T=300$}\\ 
	\cline{2-4} \cline{5-7} \cline{8-10} \cline{11-13} \cline{14-16}
	\noalign{\smallskip}
	 & {$f_{best}$} & {$f_{avg}$} & {$t_{avg}$} & {$f_{best}$} & {$f_{avg}$} & {$t_{avg}$} & {$f_{best}$} & {$f_{avg}$} & {$t_{avg}$} & {$f_{best}$} & {$f_{avg}$} & {$t_{avg}$}& {$f_{best}$} & {$f_{avg}$} & {$t_{avg}$}\\ \hline
	500-30-0 &  \textbf{116014} & \textbf{115986.8} & \textbf{3519.6} &\textbf{116014}  & 115983.7 & 3978.3 & \textbf{116014} & 115972.2  &5114.7 & \textbf{116014} & 115952.9 & 5024.7 &115864&115711.3& 6449.6\\ 
	
	500-30-1 &  \textbf{114810} & 114752.9 & \textbf{3319.3} &\textbf{114810} & \textbf{114767.3}  & 3818.5 & \textbf{114810} & 114766.8  & 4194.1 & \textbf{114810}& 114742.9& 4867.1&114705&114576.8& 6418.5\\ 
	
	500-30-2 &  \textbf{116741} & 116692.7 & 2998.7 &\textbf{116741} & 116689.5 & \textbf{2915.7} & \textbf{116741} & \textbf{116693.1}  & 4735.8 & \textbf{116741}&116677.9& 4583.5&116638&116498.8&6204.8 \\
	
	500-30-3  & \underline{\textbf{115370}} & 115292.3 & \textbf{2942.5} & \underline{\textbf{115370}} & 115291.5 & 4022.8& 115318 & \textbf{115295.0}  & 4169.7 &115321&115293.5&  5013.1&115241& 115129.3& 6410.5\\
	
	500-30-4   & \underline{116539} &  116488.6 & \textbf{3306.8} &\underline{116539} & \textbf{116493.5} & 4285.1& \underline{\textbf{116553}} & 116486.9  & 4665.3 & \underline{\textbf{116553}}& 116482.9& 4936.1&116395& 116300.5& 6325.3 \\
	
	500-30-5  & \textbf{115741} & \textbf{115736.8} & \textbf{3029.3} & \textbf{115741} &115736.3 & 3610.7& \textbf{115741} & 115734.3  & 4674.3 & \textbf{115741} &115719.7&  5824.5&115542& 115424.7& 6474.7 \\ 
	
	500-30-6  & \textbf{114181} & \textbf{114148.2} & \textbf{3223.8} & \textbf{114181} &114133.4  & 3556.5& \textbf{114181} & 114108.5  & 5426.0 &\textbf{ 114181}& 114097.9& 5403.7 &114001&113899.8&6677.2\\ 
	
	500-30-7  & \textbf{114344} &  114313.0 & 3926.5& \textbf{114344} & \textbf{114314.4} & \textbf{3800.0} & \textbf{114344} & 114311.4  &  4794.3 & \textbf{114344}& 114298.3& 5084.4&114183& 114063.9& 6799.2\\ 
	
	500-30-8  & \textbf{115419} & \textbf{115419.0} &  \textbf{578.4} & \textbf{115419} & \textbf{115419.0} & 996.0& \textbf{115419} & \textbf{115419.0}  & 2825.2 & \textbf{115419}& 115404.3& 5125.1& 115206& 115119.5& 6044.9\\
	
	500-30-9  & \textbf{117116} & 117106.8 &   \textbf{1938.9} & \textbf{117116} & 117105.7 & 2210.8& \textbf{117116} & \textbf{117110.4}  & 3957.0 & \textbf{117116}& 117107.3&  5008.7&116935& 116850.4& 6239.6 \\ 
	
	500-30-10 & \textbf{218104} & \textbf{218078.1} &  2852.8 & \textbf{218104} & 218076.1 & \textbf{2601.1} & \textbf{218104} &  218076.8  & 4335.5 & \textbf{218104}&  218066.0& 4945.6&218064& 217971.6& 6077.3 \\ 
	
	500-30-11  & 214645 & 214631.2 & \textbf{1201.9} & 214645 & 214631.7 & 1808.9& 214645 & 214636.8  & 2375.3 &\underline{\textbf{214660}}&\textbf{214640.3}& 4089.5&214585&214510.6&5611.9\\ 
	
	500-30-12  & 215945 & 215925.7 &  4001.3 & 215945 &215925.2 & \textbf{3830.4}& \textbf{215946} &\textbf{ 215926.5} &  4226.6 & 215942&  215920.4&  4380.3&215931&215871.8&6313.3 \\ 
	
	500-30-13  & \textbf{217910} & 217864.6 & \textbf{1574.9} & \textbf{217910} & 217865.6  & 2595.8&\textbf{ 217910} & \textbf{217868.1}  & 3761.4 & \textbf{217910}&  217864.1& 5196.3&217778&217694.7&6205.9 \\ 
	
	500-30-14  & 215680 &215637.0 & \textbf{2772.9}& 215640 & 215636.0  & 3199.5 & 215680 & \textbf{215638.9}  & 4105.6 & \textbf{215689}& 215632.2& 5583.5&215604&215527.5&5767.2\\ 
		
	500-30-15  & \textbf{215919} & 215873.6 & \textbf{3592.4} & \textbf{215919} &215877.5 & 3803.2& \textbf{215919} & \textbf{215881.7}  & 4460.6 &\textbf{215919}& 215864.4&5788.6 &215826&215670.7&6499.0\\ 
	
	500-30-16  & \textbf{215907} &  215884.3 & \textbf{1024.6} & \textbf{215907} &215884.3 & 1109.1& \textbf{215907} &  \textbf{215884.6 } & 2420.9 & \textbf{215907}& 215883.4& 4553.4&215773&215695.5&6143.2\\ 
	
	500-30-17 & \textbf{216542} &  216483.1 & 3444.5 & \textbf{216542}&216480.0  & \textbf{3363.9} &\textbf{ 216542} & \textbf{216496.7} &  4583.5 &216450& 216336.2& 5136.2&216371&216279.5&6138.0 \\ 
	
	500-30-18 & 217340 & \textbf{217338.2} &  \textbf{4143.9}& \underline{\textbf{217364}} & 217336.6 & 4307.9  & \underline{\textbf{217364}} &  217332.7  & 5052.3 & 217340& 217315.6&  5164.0&217231&217124.4&6170.3 \\ 
	
	500-30-19  & \textbf{214739} & 214710.4 & 3980.6& \textbf{214739} &214704.2  & \textbf{3834.4}& \textbf{214739} &  \textbf{214710.9} &  4847.5 &\textbf{214739}&214699.7&4742.7&214670&214553.7&6598.2 \\ 
	
	500-30-20  & \textbf{301675} & 301665.1 & \textbf{2769.5} & \textbf{301675} & 301666.0  & 3227.9& \textbf{301675} & \textbf{301669.4}  &3427.9 & \textbf{301675}&301666.8& 4858.3&301643&301591.8&5618.6 \\ 
		
	500-30-21  & \textbf{300055} &  300050.8 & 2849.2 & \textbf{300055} &300052.5 & \textbf{2272.7}& \textbf{300055} & 300054.2 &  3337.9 & \textbf{300055}& \textbf{300054.8}&3373.7&\textbf{300055}&300022.2&5305.6 \\
	
	500-30-22  & \textbf{305087} & 305068.0 & 3347.3& \textbf{305087} &305067.0& \textbf{3034.7} & \textbf{305087} & \textbf{305069.8 } &4300.9 & \textbf{305087}&305069.7& 5224.8&\textbf{305087}&305013.3&5486.9\\ 
	
	500-30-23  & 302008 &  302004.4 & \textbf{2396.8}& \textbf{302032} &\textbf{302006.2} & 2839.7 & \textbf{302032} & 302004.9 &4390.8 &\textbf{302032}&302004.0& 5217.6&302003&301957.5&5603.8 \\ 
	
	500-30-24 &304447 & \textbf{304431.2} & 4096.6& 304447&304425.1  & 3672.7  & 304447 & 304426.4 & \textbf{3463.5}& \underline{\textbf{304450}}&304424.4& 4046.7&304425&304375.7&6032.3 \\
	
	500-30-25  & \textbf{297012} & 296977.9 &  \textbf{3297.4} & \textbf{297012} &296979.6  & 3537.1& \textbf{297012} &  \textbf{296984.0} &  4008.9 & \textbf{ 297012}&296980.1&3409.1&297007&296968.9&4014.0 \\ 
	
	500-30-26  & \textbf{303364} & 303335.7 & \textbf{3074.1} & \textbf{303364}&303332.0  & 3626.4& \textbf{303364} & 303339.0 &3265.3 & \textbf{303364}& \textbf{303347.2}&3389.3&\textbf{303364}&303338.2&4156.8\\ 
	
	500-30-27  & 306999 & 306991.3 & \textbf{2261.3} & \textbf{307007} &\textbf{306994.5}& 3111.5& \textbf{307007} & 306991.6  &   4249.2 & 306999&306981.1& 4980.1&306976&306901.2&5755.8 \\ 
	
	500-30-28  & \textbf{303199} & \textbf{303179.0} &  \textbf{2795.3 }& \textbf{303199} &303177.5  & 2876.6& \textbf{303199} &  303175.0 & 3393.7 &\textbf{303199}&303178.5&  4323.5&303178&303138.1&5119.1 \\ 
	
	500-30-29 & \textbf{300572 }& \textbf{300537.6} & 3473.8 & \textbf{300572} &300536.4 & \textbf{3362.7 }& 300543 & 300535.5 & 4068.4 &\textbf{300572}&  300533.7&  4575.9&300555&300495.2&5324.4\\ \hline
	
	Avg.  & 211447.5 & \textbf{211420.1} & 2924.5 & \textbf{211448.0}  & 211419.6 & 3173.7& 211447.4 & 211420.0 & 4086.1 & 211447.9 &211412.9& 4795.6&211361.2&211275.9& 5932.9\\
	\#Best  & 22 & 9 & \textbf{20} & \textbf{25} & 6 & 9& \textbf{25} & \textbf{14}  & 1 & \textbf{25} & 3&0&3&0&0 \\ \hline	

\end{tabular*}
\caption{\label{tab:T}Testing parameter \textit{T} on the (500, 30)  OR-Library instances.}
\end{table*}

All the parameters of \emph{FEPSS} used in the experiments are listed in Table \ref{tab:par}. Except for the three parameters examined above, the other parameters are arbitrarily set with intuition.

\begin{table}[!h]
	\centering
\begin{tabular}{llll}
		\hline
		Parameters & Descriptions & Values & Used in \\ \hline

         \multirow{4}*{total time limit} &  \multirow{4}*{termination condition of Algorithm \ref{alg:frame}} & 360s, if $n=100$ & Algorithm \ref{alg:frame}\\
         &  & 3600s, if $n\leq250$\\
         &  & 7200s, if $n=500$\\
         &  & 36000s, if $n>500$\\
    	\hline

        \multirow{2}*{$T$} &  \multirow{2}*{time limit for each run of Integer Programming} &  10s, if $n \le 500$& Algorithm \ref{alg:ip}\\
        && 20s, if $n > 500$ &\\
        	\hline

        $ns$ & number of solutions in the solution set $SS$ & 100 & Algorithm \ref{alg:init}\\ 
        	\hline
        $maxIteration$ & number of iterations for the tabu search& 500&  Algorithm \ref{alg:tabusearch}\\ 
        	\hline
        $\alpha$ &parameter used in scoring function $V(s)$ & 0.7& Algorithm \ref{alg:frame}\\ 
        	\hline
        $\beta$ &parameter used in scoring function $\delta_i$ & 0.6& Algorithm \ref{alg:fsss}\\ 
        	\hline
        $P_1$ &probability used in  \emph{randomFunction} & 0.1& Algorithm \ref{alg:cbs}\\ 
        	\hline
        $P_2$ &probability used in  \emph{ratioFunction} & 0.7& Algorithm \ref{alg:cbs}\\ 
                	\hline
        
        \multirow{2}*{$t_1$} &number of items added into the free part in & 10& Algorithm \ref{alg:ip}\\
        & \emph{explore} procedure & \\
        	\hline
        $t_2$ &number of items used in \emph{ratioFunction} & $\frac{n}{50}$& Algorithm \ref{alg:cbs}\\
        	\hline
        $\Delta$ &parameter for the size of free part & $\frac{n}{8}$& Algorithm \ref{alg:fsss} \\
	\hline
        $lnsLimit$ & frequency of execution of \emph{LNS} & 100& Algorithm \ref{alg:frame}\\ 
        	\hline
        $lnsIterationNum$ &number of iterations in \emph{LNS}  & 10& Algorithm \ref{alg:ip}\\
		\hline
	\end{tabular}\hspace*{10pt}
	
	\caption{Parameters of FEPSS in the experiments.}
	\label{tab:par}

\end{table}

\subsection{Comparison with The Baselines}

Firstly, we present the average results of the OR-Library instances in Table \ref{tab:ave}. The results are the average of 30 instances of each combination ($n, m$). The OP/BK presents the optimal results or the best-known results.  It is shown that \emph{FEPSS} gets the best performance in the best objective comparison in 8 sets of instances, and gets the best performance in the worst objective comparison in 5 sets of instances, and gets the best performance in the average objective comparison in 6 sets of instances.  Note that it outperforms all the others in the three sets of large instances with 500 items.  Although it requires more computational time than existing algorithms for smaller instances, e.g., item numbers 100 and 250, it consumes less computational time than \emph{TPTEA} for larger instances, e.g., item number 500.

\begin{table*}[!htb]
\centering
\scriptsize\setlength{\tabcolsep}{0.5pt}
\begin{tabular*}{\linewidth}{c@{\extracolsep{\fill}}c*{14}{c} }
\hline
Instances& &\multicolumn{4}{c}{$f_{best}$} 
&\multicolumn{3}{c}{$f_{worst}$}
&\multicolumn{3}{c}{$f_{avg}$}
&\multicolumn{3}{c}{$t_{avg}(s)$}\\ 
	\cline{3-6} \cline{7-9} \cline{10-12} \cline{13-15}
	\noalign{\smallskip}
	 ($n, m$) & OP/BK & \emph{CPlex} &\emph{DQPSO} & \emph{TPTEA} & \emph{FEPSS}
	& \emph{DQPSO} & \emph{TPTEA} & \emph{FEPSS}  &  \emph{DQPSO} & \emph{TPTEA} & \emph{FEPSS} & \emph{DQPSO} & \emph{TPTEA} & \emph{FEPSS} \\ \hline
	(100, 5) & 42640.4  &\textbf{42640.4} & \textbf{42640.4} & \textbf{42640.4} & \textbf{42640.4} &42640.0&42640.0&\textbf{42640.1}& 42640.0 & 42640.2 & \textbf{42640.3}  &\textbf{0.4} & 5.5 & 17.6 \\
	
	(100, 10) & 41606.0  & \textbf{41606.0} & \textbf{41606.0} & \textbf{41606.0} & \textbf{41606.0} &41603.0&\textbf{41603.5}&41591.8& \textbf{41604.7} & 41604.5 & 41602.8  &\textbf{2.4} & 8.1 & 68.7 \\
	
	(100, 30) & 40767.5 &\textbf{40767.5} & 40765.3 & 40766.5 & 40764.4&40757.7&\textbf{40759.1}&40727.6 & \textbf{40763.4} & 40760.4 & 40754.7  &\textbf{7.7} & 9.9 & 93.2 \\
	
	(250, 5) & 107088.9  & \textbf{107088.9} &  \textbf{107088.9} & \textbf{107088.9} & \textbf{107088.9} &107088.7&\textbf{107088.8}&\textbf{107088.8 }& 107088.8 & 107088.4 & \textbf{107088.9} & \textbf{71.9} & 260.0 & 281.7 \\ 
	
	(250, 10) & 106365.7  & 106364.8& \textbf{106365.7} & \textbf{106365.7} & \textbf{106365.7} &106359.8&\textbf{106362.5}&106362.2& 106363.3 & 106362.6 & \textbf{106365.4}  & \textbf{333.2} & 350.5 & 688.5 \\ 
	
	(250, 30) & 104717.7  & 104696.0 & 104706.9 & 104717.0 & \textbf{104717.2} &104679.1&\textbf{104704.7}&104698.0& 104695.2 & \textbf{104713.0} & 104708.8 & 686.5 & \textbf{349.7} & 1191.5 \\

	(500, 5) & 214168.8 &214168.0 & 214168.4 & 214167.3 & \textbf{214168.8} &214166.3&214150.0&\textbf{214166.9}& 214167.4 & 214160.6 & \textbf{214168.4} &\textbf{810.7} & 2650.7 & {1383.7}
    \\ 
	(500, 10) & 212859.3  & 212832.8& 212850.9 & 212841.4 & \textbf{212856.2} &212823.2&212784.1&\textbf{212832.7}& 212837.6 & 212807.0 & \textbf{212844.0} & 2473.6 & 3457.6 & \textbf{2446.3}
    \\ 
	(500, 30) & 211453.7  & 211374.2& 211399.2 & 211421.8 & \textbf{211448.0} &211270.7&211339.4&\textbf{211396.5}& 211353.1 & 211371.7 & \textbf{211419.6}  & \textbf{2777.0} & 3638.9 & 3173.7 \\ \hline
	Avg. & 120185.3 & 210170.9 & 120176.9 & 120179.4 & \textbf{120184.0} &120154.3&120158.0&\textbf{120167.2}& 120168.2 & 120167.6 & \textbf{120177.0}  & \textbf{795.9} & 1192.3 & 1038.3 \\ 
	\#Best & & 4 & 4 &4 &\textbf{8} &0&\textbf{5}&\textbf{5}& 2 & 1 & \textbf{6}  &\textbf{7} & 1 & 1 \\ \hline

\end{tabular*}
\caption{\label{tab:ave}Average results of OR-Library instances.}
\end{table*}

Secondly, we present the detailed results of the largest OR-Library instances (500-30).  The BK column represents the best-known objectives of the instances. We include the BK column in the $f_{best}$ comparison. It is shown that our \emph{FEPSS} finds the best-known objectives for 23 instances,  while also establishing new lower bounds for 3 instances (the underlined results): 500.30.03, 500.30.04, and 500.30.18. In addition, From the $f_{avg}$ columns, we observe \emph{FEPSS} achieves the best performance in 28 instances.  Although \emph{DQPSO} costs less time than the other two, it is outperformed by the others in solution quality. The column $f_{worst}$ shows the worst case of the 30 runs of each algorithm. We can see that \emph{FEPSS} achieves the best performance in all 30 instances in the $f_{worst}$ column. Furthermore, from the average of the standard deviations ($Std$ columns), we can see that our algorithm has the lowest value, which indicates that \emph{FEPSS} is more stable than the other algorithms. The small p-values ($<0.05$) from the Wilcoxon signed-rank tests indicate that the results of \emph{FEPSS} significantly differ from those of other algorithms in terms of $f_{best}$.

Finally, we present the results of MK\_GK instances in Table \ref{tab:mk}. This set contains some large instances, e.g., n=2500 and m=100.  We can see that our  \emph{FEPSS} finds new lower bounds for the largest 5 instances. It finds the best solutions for all the 11 instances. It also gets the best performance in the average objective of the 30 runs in 10 instances. Besides, it takes less time to find the best-found solutions in the largest instances than the others. From $f_{worst}$, we can observe that our algorithm wins in 9 instances out of 11, along with the smallest average standard deviation indicating the stability of \emph{FEPSS}.

\begin{landscape}
\begin{table*}[!htb]
\centering
\scriptsize\setlength{\tabcolsep}{0.5pt}
\begin{tabular*}{\linewidth}{c@{\extracolsep{\fill}}c*{21}{c} }
\hline
\multirow{2}*{Instance}& \multicolumn{5}{c}{$f_{best}$} 
&\multicolumn{4}{c}{$f_{worst}$}
&\multicolumn{4}{c}{$f_{avg}$}
&\multicolumn{4}{c}{$Std.$}
&\multicolumn{4}{c}{$t_{avg}(s)$}
\\ 
	\cline{2-6} \cline{7-10} \cline{11-14}\cline{15-18}\cline{19-22}
	\noalign{\smallskip}
	 & BK  & \emph{CPlex} & \emph{DQPSO} & \emph{TPTEA} & \emph{FEPSS}
	&\emph{CPlex}& \emph{DQPSO} & \emph{TPTEA} & \emph{FEPSS} &\emph{CPlex}& \emph{DQPSO} & \emph{TPTEA} & \emph{FEPSS} &\emph{CPlex}& \emph{DQPSO} & \emph{TPTEA} & \emph{FEPSS} & \emph{CPlex}&\emph{DQPSO} & \emph{TPTEA} & \emph{FEPSS}\\ \hline
	500-30-0 & \textbf{116056}  & 116014 & 115952 & 115957 & 116014 &115903&115746&115876&\textbf{115924}&115956.5& 115883.8 & 115912.4& \textbf{115983.7} &28.8& 62.4&  19.0& 33.7&6278.0 &\textbf{2359.1} & 3295.2 & 3978.3 \\ 
	
	500-30-1 & \textbf{114810} & \textbf{114810}  & 114734 & 114769 & \textbf{114810}&114718 &114571&114689&\textbf{114732}&114763.5& 114711.2 & 114722.6 & \textbf{114767.3}  & 32.1& 14.1& 23.4& 31.3&6571.6& \textbf{3455.2} & 3522.6 & 3818.5 \\ 
	
	500-30-2 & \textbf{116741} & 116712 & 116712 & 116682 & \textbf{116741} &116628&116455&116569&\textbf{116661}&116682.9& 116629.7 & 116614.1 & \textbf{116689.5}& 23.1 & 35.3& 29.0& 25.5&6460.1  & \textbf{2027.1} & 3914.3 & 2915.7\\
	
	500-30-3 & {115354} & 115337 & 115258 & 115313 & \underline{\textbf{115370}}&115228&115182&115202&\textbf{115264} &115279.3& 115238.7 & 115247.2 & \textbf{115291.5}  & 32.5& 9.3& 26.9& 29.2& 5922.0&\textbf{2855.1} & 3685.6 & 4022.8 \\
	
	500-30-4  &{116525} & 116516 & 116437 & 116471 & \underline{\textbf{116539}} &116419&116287&116344&\textbf{116432}&116461.3& 116362.0 & 116392.2 & \textbf{116493.5} & 25.6& 29.0& 30.8& 26.6&6683.7 & \textbf{3851.5} & 3870.8 & 4285.1\\
	
	500-30-5 & \textbf{115741} &\textbf{115741} & 115701 & 115734 & \textbf{115741}&115642&115598&115622& \textbf{115734}& 115722.1&115655.8 & 115671.7 & \textbf{115736.3}& 21.7& 13.8& 27.2& 3.3& 5703.6 &\textbf{1566.2} & 4044.7 & 3610.7 \\ 
	
	500-30-6 & \textbf{114181} &114148  & 114085 & 114111 & \textbf{114181}&114007&113915&113969&\textbf{114072} &114067.0& 113991.1 & 114017.0 & \textbf{114133.4} & 40.8 & 55.4& 32.0& 40.6&5896.9 &3607.2 & 4255.9 & \textbf{3556.5}\\ 
	
	500-30-7 & \textbf{114403} & 114314 & 114243 & 114248 & 114344 &114180&113945&114128&\textbf{114261}&114255.2& 114158.9 & 114172.1 & \textbf{114314.4}  & 35.7& 25.7& 27.7& 26.0& 6105.5 &\textbf{3707.9} & 4389.8 & 3800.0 \\ 
	
	500-30-8 & \textbf{115419} & \textbf{115419} & \textbf{115419} & 115282 & \textbf{115419} &115225&115149&115178&\textbf{115419}&115408.1& 115257.9 & 115282.2 & \textbf{115419.0} & 41.4&64.4& 23.6& 0.0&6118.9 & 3980.0 & 4113.6 & \textbf{996.0} \\
	
	500-30-9 & \textbf{117116} & \textbf{117116} & 117030 & 117104 & \textbf{117116} &117006&116868&116944&\textbf{117058}&117064.0& 116989.5 & 116988.1 & \textbf{117105.7}& 43.0& 28.2 & 33.6& 10.3 &5931.1 & 3636.4 & 3891.4 & \textbf{2210.8}\\ 
	
	500-30-10 & \textbf{218104} & \textbf{218104} & 218053 & \textbf{218104} & \textbf{218104} &218008&217979&218051&\textbf{218068}&218059.5& 218041.7 & 218075.2 & \textbf{218076.1}  & 22.3 & 6.4& 11.9& 10.8&6160.0& \textbf{1509.6} & 2965.0 & 2601.1\\ 
	
	500-30-11 & \textbf{214648} & 214629 & 214626 & 214645 & 214645 &214533&214428&214519&\textbf{214626}&214594.0 &214530.9 & 214565.2 & \textbf{214631.7} & 36.9& 27.8& 37.4 & 8.7&5848.7 &3364.7 & 3810.9 & \textbf{1808.9}\\ 
	
	500-30-12 & \textbf{215978} &215918 & 215905 & 215945 & 215945 &215842&215854&215867&\textbf{215905}&215883.0& 215877.4 & 215901.4 & \textbf{215925.2}  & 24.4& 10.7 & 20.7& 11.8&5717.3&\textbf{2623.4} & 3610.3 & 3830.4\\ 
	
	500-30-13 & \textbf{217910} & 217892  & 217825 & \textbf{217910} & \textbf{217910} &217781&217754&217775&\textbf{217862}&217843.4& 217798.5 & 217817.2 & \textbf{217865.6}  & 32.4 & 17.2& 26.6& 10.6&5430.0&2835.4 & 3987.7 & \textbf{2595.8}\\ 
	
	500-30-14 & \textbf{215689} & 215640  & 215649 & 215640 & 215640 &215571&215440&215609&\textbf{215613}& 215610.2&215591.7 & 215618.0 & \textbf{215877.5} & 20.5 & 23.3& 10.9& 8.0&5381.2 & \textbf{2079.1} & 3213.3 & 3803.2\\ 
		
	500-30-15 & \textbf{215919} &215867 & 215832 & 215825 & \textbf{215919} &215789&215662&215720&\textbf{215853}&215826.3& 215775.5 & 215758.1 & \textbf{215877.5}  & 20.0 & 23.7& 22.5 & 21.5&5535.4 & \textbf{3262.4} & 4537.5 & 3803.2\\ 
	
	500-30-16 & \textbf{215907} &\textbf{215907} & 215883 & \textbf{215907} & \textbf{215907} &215788&215783&215826&\textbf{215883}& 215839.2&215822.6 & 215848.2 & \textbf{215884.3}& 36.9& 44.5 & 13.1& 5.1 & 5755.7& 3422.1 & 2812.8 & \textbf{1109.1}\\ 
	
	500-30-17 & \textbf{216542}&216510  & 216450 & \textbf{216542} & \textbf{216542} &216419&216342&216419&\textbf{216439}&216439.1& 216407.6 & 216461.5 & \textbf{216480.0} & 17.0 & 36.8& 29.5& 32.5&5889.0& \textbf{3085.3} & 3894.9 & 3363.9\\ 
	
	500-30-18 & {217340} & \underline{\textbf{217364}} & 217329 & 217337 & \underline{\textbf{217364}} &217234&217234&217291&\textbf{217313}&217320.4& 217271.7 & 217313.5 & \textbf{217336.6} & 29.0& 15.1  & 10.3& 10.4&5887.0 &3313.1 & \textbf{3153.9} & 4307.9\\ 
	
	500-30-19 & \textbf{214739} &\textbf{214739}  & 214681 & 214701 & \textbf{214739} &214633&214596&214639&\textbf{214682}&214678.4& 214664.6 & 214668.2 & \textbf{217704.2} & 24.0 & 21.7& 15.5& 22.0&6121.1 & \textbf{2332.8} & 4278.1 & 3834.4\\ 
	
	500-30-20 & \textbf{301675} & \textbf{301675}  & 301643 & \textbf{301675} & \textbf{301675} &301643&301623&301637&\textbf{301656}&301650.0& 301642.3 & 301646.1 & \textbf{301666.0}  & 11.0& 3.6& 8.9& 7.8& 6213.0&\textbf{629.0} & 3005.8 & 3227.9\\ 
		
	500-30-21 & \textbf{300055} &\textbf{300055} & \textbf{300055} & \textbf{300055} & \textbf{300055} &299958&299919&300009&\textbf{300030}&300007.4& 299997.3 & 300038.4 & \textbf{300052.5} & 27.3 & 23.1& 18.8& 5.1&6181.7& 2845.2 & 3317.3 & \textbf{2272.7}\\
	
	500-30-22 & \textbf{305087} &305080& 305055 & \textbf{305087} & \textbf{305087} &305002&304962&\textbf{305055}&\textbf{305055}&305041.6& 305044.6 & \textbf{305080.2} & 305067.0 & 18.9 & 6.7& 9.0& 10.7&6414.3 &3749.1 & 3209.9 & \textbf{3034.7}\\ 
	
	500-30-23 & \textbf{302032} &302021 & 301988 & \textbf{302032} & \textbf{302032} &301961&301935&301965&\textbf{302004}&301990.0& 301939.1 & 301989.5 & \textbf{302006.2}& 16.5& 10.0& 17.3& 7.0&6236.4 &\textbf{1178.3} & 3507.4 & 2839.7 \\ 
	
	500-30-24 & \textbf{304462}&\textbf{304462}  & 304423 & \textbf{304462} & 304447 &304398& 304386 &304406&\textbf{304413}&304423.4& 304401.5&304423.7 & \textbf{304425.1}& 14.2& 7.4& 9.0& 9.4& 6043.5 &\textbf{965.9}& 3655.4 & 3672.7\\
	
	500-30-25 & \textbf{297012} &\textbf{297012}  & 296962 & 296998 & \textbf{297012} &296892&296832&296942&\textbf{296959}&296940.9 &296947.5 & 296963.9 & \textbf{296979.6} & 24.2& 20.7& 13.4& 19.2&6388.5&\textbf{2152.4} & 3670.3 & 3537.1 \\ 
	
	500-30-26 & \textbf{303364}&303335  & 303360 & \textbf{303364} & \textbf{303364} &303276&303201&303323&\textbf{303322}&303310.1& 303324.7 & \textbf{303340.0} & 303332.0 & 16.8& 26.9 & 14.8& 10.7&6495.8&4024.2 & \textbf{3388.6} & 3626.4\\ 
	
	500-30-27 & \textbf{307007} &306999& 306999 & 306999 & \textbf{307007} &306930&306893&306945&\textbf{306961}&306957.5 &306963.9 & 306973.1 & \textbf{306994.5} & 20.0& 26.8 & 12.9& 12.6&6314.9&\textbf{2566.3} & 3581.5 & 3111.5 \\ 
	
	500-30-28 & \textbf{303199} &\textbf{303199}  & 303162 & \textbf{303199} & \textbf{303199} &303107&303100&303158&\textbf{303162}&303156.8& 303148.7 & 303170.1 & \textbf{303177.5} & 23.1 & 10.8& 13.5 & 9.1& 6482.7&3650.6 & 3373.2 & \textbf{2876.6}\\ 
	
	500-30-29 & \textbf{300596} &300572 &300536 & 300572 & 300572 &300506&300481&300512&\textbf{300532}&300533.5& 300532.0 & 300532.4 & \textbf{300536.4} & 11.0& 12.1& 12.0& 7.1&6080.3& 3873.2 &\textbf{3209.1} & 3362.7 \\ \hline
	
	Avg. & \textbf{211451.9}  & 211438.0 & 211399.2 & 211422.3 & {211448.0} &211340.9&211314.8&211339.6&\textbf{211396.5}&211392.1& 211353.1 & 211371.6 & \textbf{211419.6} & 25.7& 23.8&20.0&15.6& 6074.9& \textbf{2777.0} & 3638.9 & 3173.7\\
	\#Best & \textbf{27} & 13 & 2 & 11 & {23} &0&0&1&\textbf{30} & 0&0 & 2 & \textbf{28} &&&&&0& \textbf{18} & 3 & 9 \\ 
	p-value& 1.02e-1 & 1.45e-3 & 5.23e-6 &1.12e-3\\	\hline

\end{tabular*}
\caption{\label{tab:500-30}Detailed results of the (500, 30)  OR-Library instances.}
\end{table*}

\begin{table*}[!htb]
\centering
\scriptsize\setlength{\tabcolsep}{0.5pt}
\begin{tabular*}{\linewidth}{c@{\extracolsep{\fill}}c*{22}{c} }
\hline
\multirow{2}*{Instance}&\multirow{2}*{ ($n, m$)} &  \multicolumn{5}{c}{$f_{best}$} 
&\multicolumn{4}{c}{$f_{worst}$}
&\multicolumn{4}{c}{$f_{avg}$}
&\multicolumn{4}{c}{$Std.$}
&\multicolumn{4}{c}{$t_{avg}(s)$}
\\ 
	\cline{3-7} \cline{8-11} \cline{12-15} \cline{16-19}\cline{20-23}
	\noalign{\smallskip}
	 &  & BK  & \emph{CPlex} & \emph{DQPSO} & \emph{TPTEA} &  \emph{FEPSS} & \emph{CPlex}&  \emph{DQPSO} & \emph{TPTEA} &  \emph{FEPSS}  & \emph{CPlex}& \emph{DQPSO} & \emph{TPTEA} &  \emph{FEPSS} & \emph{CPlex}& \emph{DQPSO} & \emph{TPTEA} &  \emph{FEPSS} & \emph{CPlex}& \emph{DQPSO} & \emph{TPTEA} &  \emph{FEPSS}\\ \hline
	mk\_gk01 &(100, 15)&\textbf{3766}&\textbf{3766}&\textbf{3766}&\textbf{3766}&\textbf{3766}&\textbf{3766}&3763&\textbf{3766}&\textbf{3766}&\textbf{3766.0}&3765.2&\textbf{3766.0}&\textbf{3766.0}&0.0&1.3&0.0&0.0&\textbf{5.0}&61.9&34.4&68.6 \\
	
	mk\_gk02 &(100, 25)&\textbf{3958}&\textbf{3958}&\textbf{3958}&\textbf{3958}&\textbf{3958}&3796&3950&3956&\textbf{3958}&3899.9&3956.0&3956.7&\textbf{3958.0}&48.4&1.9&0.8&0.0&131.5&96.6&\textbf{66.48}&96.6 \\
	
	mk\_gk03 &(150, 25)&\textbf{5656}&\textbf{5656}&5652&\textbf{5656}&\textbf{5656}&\textbf{5652}&5648&5650&\textbf{5652}&5655.4&5650.2&\textbf{5655.5}&5653.9&0.8&0.9&1.6&1.7&2849.4&\textbf{741.1}&1388.3&1695.2 \\
	
	mk\_gk04 &(150, 50)&\textbf{5767}&\textbf{5767}&5764&\textbf{5767}&\textbf{5767}&5765&5759&\textbf{5767}&\textbf{5767}&5766.7&5762.8&\textbf{5767.0}&\textbf{5767.0}&0.7&1.5&0.0&0.0&2852.1&\textbf{792.1}&865.9&1217.8 \\
	
	mk\_gk05 &(200, 25)&\textbf{7561}&\textbf{7561}&7560&\textbf{7561}&\textbf{7561}&7506&7553&\textbf{7559}&\textbf{7559}&7556.2&7556.0&7559.8&\textbf{7560.1}&13.0&1.2&0.6&0.4&2851.9&\textbf{912.7}&1546.2&1415.7 \\
	
	mk\_gk06 &(200, 50)&\textbf{7680}&7679&7672&7678&\textbf{7680}&7673&7665&\textbf{7675}&7674&7675.9&7668.7&7676.2&\textbf{7676.9}&1.5&1.7&0.7&1.1&3159.1&\textbf{969.5}&1330.8&2089.1 \\
	
	mk\_gk07 &(500, 25)&{19220}&19220&19215&19219&\underline{\textbf{19221}}&19054&19210&19215&\textbf{19219}&19205.5&19211.8&19216.5&\textbf{19219.2}&36.6&1.2&1.0&0.5&6549.1&\textbf{3160.2}&3834.4&3305.0 \\
	
	mk\_gk08 &(500, 50)&{18806}&18806&18792&18804&\underline{\textbf{18808}}&18801&18779&18799&\textbf{18803}&18803.6&18784.3&18800.4&\textbf{18805.0}&1.4&3.0&1.2&1.2&5703.5&5015.0&3900.7&\textbf{3879.1} \\
	
	mk\_gk09 &(1500, 25)&{58087}&58089&58085&58089&\underline{\textbf{58091}}&58019&58076&58084&\textbf{58088}&58085.8&58080.0&58086.0&\textbf{58089.1}&12.4&2.0&1.1&0.7&33183.3&\textbf{13606.6}&25661.3&16514.5 \\
	
	mk\_gk10 &(1500, 50)&{57295}&57292&57274&57291&\underline{\textbf{57296}}&57239&57258&57289&\textbf{57292}&57287.7&57265.4&57289.9&\textbf{57293.1}&9.1&4.2&1.0&1.1&32407.8&28035.2&25438.7&\textbf{17846.0} \\
	
	mk\_gk11 &(2500, 100)&{95237}&95234&95179&95233&\underline{\textbf{95239}}&95225&95156&95222&\textbf{95231}&95229.0&95168.2&95227.2&\textbf{95232.8}&2.1&5.0&2.6&1.6&27963.1&34592.9&25669.5&\textbf{21757.4} \\ \hline
	Avg. & &25730.3&25729.8&25719.7 &25729.3 &\textbf{25731.2} &25681.5&25710.6&25725.6&\textbf{25728.1}&25721.1&25715.3 & 25727.4 & \textbf{25729.2}&11.5&2.2&1.0&0.8&10696.0&7998.5&8157.8 &\textbf{6353.2} \\
	\#Best & & 6&5 &2&5&\textbf{11}&2&1&4&\textbf{9}&1&0&3&\textbf{10}&&&&&1&\textbf{6}&1&3\\ 
    p-value& &4.41e-2&2.68e-2&7.63e-3&1.79e-2\\\hline
\end{tabular*}
\caption{\label{tab:mk}Detailed results of MK\_GK instances.}
\end{table*}

\end{landscape}

\section{Conclusion and future work}

In this paper, we propose a novel algorithm that combines evolutionary computation with an exact algorithm for the 0-1 Multidimensional Knapsack Problem. The algorithm takes advantage of an efficient evolutionary algorithm to find promising search spaces, and then an exact algorithm is employed to explore the search spaces. We performed extensive experimentation with commonly used benchmark sets.  The ablation study shows that the scoring function $\delta_i$ is effective in identifying the hard-to-decide items, so it is more efficient to find high-quality solutions in the promising search spaces. Consequently, our algorithm outperforms the state-of-the-art heuristics algorithms in solution quality. It finds a new lower bound for 10 hard instances. The new algorithm is more efficient in solving some hard and large 0-1 Multidimensional Knapsack Problems.

In the future, we would apply \emph{FEPSS} framework in solving other combinational optimization problems. In particular, we would focus on the problems with well-established exact algorithms and population-based evolutionary algorithms, as well as an objective function that can be calculated at a low cost. This is because the algorithm relies on a population to find good partial assignments that correspond to promising search space. If the population can not provide useful information, the algorithm will be less efficient.

\section{Acknowledgements}
This work is supported by National Natural Science Foundation of China under grant NO. 62276060, 61802056, 61976050 and Natural Science Foundation of Jilin Province under grant NO. 20210101470JC.

\bibliography{mkp}

\end{document}